\documentclass[12pt, draftclsnofoot, onecolumn]{IEEEtran}
\usepackage{bm,cite,float,amsmath,amssymb,amsthm}

\usepackage{algorithm}
\usepackage[noend]{algpseudocode}
\usepackage{indentfirst}
\usepackage{xcolor}
\usepackage{dsfont}
\usepackage{nicefrac} 

\makeatletter
\def\BState{\State\hskip-\ALG@thistlm}
\makeatother

\usepackage{amssymb}
\usepackage{amsmath}
\usepackage{graphicx}
\usepackage{cite}
\usepackage{citesort}
\usepackage{balance}
\usepackage[utf8]{inputenc}
\usepackage{makecell}

\newtheorem{theo}{Theorem}
\newtheorem{lemma}{Lemma}

\newtheorem{ppro}{Proposition}

\IEEEoverridecommandlockouts

\usepackage{amsthm}
\usepackage{graphicx,epstopdf}
\usepackage{epsfig}	
\usepackage{amsfonts,balance}
\usepackage{bbm}
\floatname{algorithm}{Algorithm}
\usepackage{lipsum}

\newtheorem{theorem}{Theorem}

\newtheorem{corollary}{Corollary}

\makeatletter
\def\ScaleIfNeeded{%
\ifdim\Gin@nat@width>\linewidth \linewidth \else \Gin@nat@width
\fi } \makeatother

\DeclareMathOperator*{\argmin}{arg\,min}
\DeclareMathOperator*{\argmax}{arg\,max}

\begin{document}

\title{Time-triggered Federated Learning \\ over Wireless Networks}

\author{Xiaokang~Zhou, Yansha~Deng,~\IEEEmembership{Member,~IEEE,} Huiyun~Xia,\\
        Shaochuan~Wu,~\IEEEmembership{Senior Member,~IEEE,}
        and~Mehdi Bennis,~\IEEEmembership{Fellow,~IEEE}.

\vspace{-0.2cm}
\thanks{This work was supported in part by the Natural Science Foundation of China under Grants 61671173, 62171163, and 61831002, and in part by Engineering and Physical Sciences Research Council (EPSRC), U.K., under Grant EP/W004348/1. The work of X. Zhou was supported by the China Scholarship Council. This work was presented in part at the 2022 IEEE International Conference on Communications \cite{xiaokang2022icc}. (Corresponding authors: Shaochuan Wu, Yansha Deng)}
\thanks{
X. Zhou, H. Xia and S. Wu are with the School of Electronics and Information Engineering, Harbin Institute of Technology, Harbin, 150001, China (emails: kangsenneo@sina.com; summerxiahy@163.com; scwu@hit.edu.cn). This work is done during X. Zhou's visit in King's College London.}
\thanks{Y. Deng is with Department of Engineering, King's College London, London, WC2R 2LS, UK (email: yansha.deng@kcl.ac.uk).}
\thanks{M. Bennis is with the Centre for Wireless Communications (CWC), University of Oulu, 90570 Oulu, Finland (email: mehdi.bennis@oulu.fi).}

}

\maketitle

\begin{abstract}
The newly emerging federated learning (FL) framework offers a new way to train machine learning models in a privacy-preserving manner. However, traditional FL algorithms are based on an event-triggered aggregation, which suffers from stragglers and communication overhead issues. To address these issues, in this paper, we present a time-triggered FL algorithm (TT-Fed) over wireless networks, which is a generalized form of classic synchronous and asynchronous FL. Taking the constrained resource and unreliable nature of wireless communication into account, we jointly study the user selection and bandwidth optimization problem to minimize the FL training loss. To solve this joint optimization problem, we provide a thorough convergence analysis for TT-Fed. Based on the obtained analytical convergence upper bound, the optimization problem is decomposed into tractable sub-problems with respect to each global aggregation round, and finally solved by our proposed online search algorithm. Simulation results show that compared to asynchronous FL (FedAsync) and FL with asynchronous user tiers (FedAT) benchmarks, our proposed TT-Fed algorithm improves the converged test accuracy by up to 12.5\% and 5\%, respectively, under highly imbalanced and non-IID data, while substantially reducing the communication overhead.
\end{abstract}

\begin{IEEEkeywords}
Federated learning, resource allocation, convergence analysis.
\end{IEEEkeywords}

\maketitle

\section{Introduction}

 
\IEEEPARstart{O}{ver} the past decades, portable smart devices with in-built high-definition sensors have gained access to an increasing amount of customized user data \cite{erricsson2021mobility}. Trained via such massive data, machine learning-based applications provide a revolutionary way to distill information from these data resources and will continue thriving \cite{sattler2020robust}. However, the training procedure often involves frequent data sharing to a data center or cloud, leading to a growing concern about the leakage of privacy-sensitive user data\cite{tianli2020federated}.


The newly emerging federated learning (FL) \cite{google2019advFL} is a promising solution for both privacy issues and resource-constrained data transmission problems \cite{mcmahan2017communicationefficient,Lim2020Federated}. In FL, the learning task is solved in an iterative way by leveraging the local computation capabilities at users to train a shared model coordinated by the \emph{server}, without sharing any user data \cite{xie2020asynchronous,howard2020scheduling,shi2021joint}. The detailed steps of FL are: 1) at the beginning of each iteration, the server distributes the current global model to the selected users; 2) the selected users perform local updates using their local data sets in parallel based on the received global model and then upload local models to the server; 3) the server aggregates these local models to generate a new global model, and the iteration goes on until convergence.


Based on how the aggregation is triggered, FL can be classified into two categories: 1) Synchronous (Sync) FL \cite{mcmahan2017communicationefficient,wang2019adaptive,xu2021client,yang2020federated}, where the global aggregation at the server is triggered until \emph{all updates from the selected users are received}\footnote{Aggregation deadline may apply to avoid useless waiting for offline users.}; 
2) Asynchronous (Async) FL \cite{xie2020asynchronous,Hyun2021adaptive,john2021async,marten2020async}, where the global aggregation at the server is triggered whenever \emph{an update from any selected user is received}. 
Fig. \ref{Fig_intro_Async_vs_Sync} (please refer to page 8) shows the work-flow of Sync FL and Async FL, where each gray bar represents a local updating round for a specific user and the dashed lines represent the timings of global aggregations at the server.

Owing to the high communication efficiency gained in synchronized training settings, Sync FL algorithms are widely studied in wireless networks \cite{yang2021energy,chen2021ajoint,vu2020cellfree,dinh2021federated,salehi2021unreliable}, and the main challenges of its wireless implementation lie in the resource-constrained and unreliable nature of wireless networks. Due to limited communication resource budgets, the user selection problem was studied to maximize the resource utilization with the focus on energy efficiency \cite{yang2021energy}, training accuracy \cite{vu2020cellfree}, training time\cite{dinh2021federated}, and etc. However, these works all assume that the uploaded models are successfully decoded at the server, and ignore the potential negative impact brought by the unreliable nature of wireless networks. Taking the practical transmission failure/error into consideration, the joint resource allocation and user selection problems were studied in \cite{chen2021ajoint,salehi2021unreliable}. 

Despite its benefits in high communication efficiency, Sync FL can result in low training efficiency, since its convergence speed is limited by the slowest users, known as stragglers \cite{vu2021straggler}. To handle the straggler issue, Async FL is a more flexible solution by allowing Async aggregation without waiting for the stragglers \cite{wang2021asynchronous,yujing2019asynchronous}. However, Async FL has its own unique challenges, including: 1) the staleness problem, where the local model based on an old global model may be harmful to the current aggregation. This problem may be solved by applying a mixing function to exclude the users with ``toxic'' local updates \cite{xie2020asynchronous};  2) the computation bias, where faster users contribute more to the global model and lead to bias. This bias issue can be solved by introducing weight factors to control the influence of different users during the global aggregation step \cite{Hyun2021adaptive}; and 3) the high communication overhead, where the frequent information exchange between the user and the server can easily induce a communication bottleneck and result in low communication efficiency. The wireless implementation of Async FL is still in its infancy, and the above works did not consider the wireless settings.

To overcome the training efficiency drawback in Sync FL and the communication efficiency drawback in Async FL, grouping users into small tiers (or clusters) and aggregating their models with respect to each tier is a promising solution. The authors in \cite{zheng2020TiFL} proposed a tier-based Sync FL, where users with similar computation delay are grouped and selected for aggregation to solve the straggler issue. Extending from \cite{zheng2020TiFL}, a multi-tier FL was first proposed in \cite{Zheng2020FedAT}, namely FedAT, based on intra-tier aggregation in a Sync way and the inter-tier aggregation in an Async way. While promising, the above studies could result in unstable and degraded performance under highly  non-IID user data, and they overlook the impact of wireless communication.


Notably, the aforementioned FL algorithms are all based on an event-triggered aggregation behavior, where the global aggregation occurs only if a specific event happens. This will lead to fluctuated global aggregation round durations, which not only causes stragglers in Sync FL, but also leads to unstable training results and high communication overhead in Async FL. Specifically, in Sync FL, since all updates from the selected users should be received before aggregation, the time duration of a global aggregation round is prolonged by waiting for stragglers, which results in low training efficiency. On the other extreme, in Async FL, since the global aggregation is triggered whenever a local update is received, those local models arrive at the server over different but similar time instances will trigger the global aggregation several times in a short duration, namely, \emph{aggregation glitch} (An example is shown in Fig. \ref{Fig_intro_Async_vs_Sync}, marked by the orange circle.). This is not desirable especially in highly non-IID data scenarios, as more local models would be required to merge into the global model to accelerate the training and to generate a more robust model. The hybrid multi-tier FedAT can alleviate the aggregation glitch to some degree, but still, it can not be solved fundamentally (As shown in Fig. \ref{Fig_intro_Async_vs_Sync}, the aggregation glitch still exists in FedAT). Moreover, due to the Async inter-tier aggregation of FedAT, broadcasting can not be leveraged to distribute global model for users in different tiers.

To solve the above issues, one promising solution is to conduct global aggregation at a fixed time interval. This allows local models from users who managed to complete their local updating within the same interval to be aggregated together. By doing so, users are naturally partitioned into different tiers, and it is possible for the global model to be broadcast to users in different tiers for communication overhead reduction, as they are aligned at the beginning of their local updating round. We coin this solution as multi-tier time-triggered FL, which reaps the benefits of both Sync and Async FL. To our best knowledge, \emph{this is the first work that proposes a multi-tier time-triggered FL framework for wireless networks}. Our major contributions are:
\begin{itemize}
\item We propose a novel multi-tier time-triggered federated learning algorithm (TT-Fed) by including existing Sync and Async FL algorithms as special cases, with the aim to achieve a good balance between training and communication efficiencies. Specifically, the users are naturally grouped into multiple small tiers to alleviate the staleness problem among users. To control the computation bias, we also consider weighting factors for different tiers and design a novel global aggregation scheme. 
\item  We implement our proposed TT-Fed algorithm over resource-constrained and unreliable wireless networks. We formulate a joint user selection and bandwidth allocation problem for training loss minimization. To quantify the convergence, we perform a thorough convergence analysis for TT-Fed to obtain the analytical convergence upper bound for the training loss function. Next, we provide a sufficient condition for parameter setting, which ensures the convergence of TT-Fed. We further analyze in detail the impact of wireless communication and global aggregation round duration on the convergence speed and accuracy.
\item  To solve the joint user selection and bandwidth allocation problem, we propose a decomposition strategy based on the above convergence analysis. Accordingly, the aforementioned optimization problem is decomposed into tractable sub-problems with respect to each global aggregation round, which is addressed by our proposed online search algorithm.
\item We evaluate the performance of TT-Fed in comparisons with three state-of-the-art FL algorithms, i.e., FedAvg, FedAsync, and FedAT, under various non-IID data and system parameter settings. Our simulation results demonstrate that under IID data setting, our proposed TT-Fed algorithm achieves similar test accuracy and convergence speed as other three existing FL algorithms; however, under highly imbalanced and non-IID data settings, our proposed TT-Fed algorithm obtains the fastest convergence speed and can improve the test accuracy by up to 12.5\% and 5\% compared to FedAsync and FedAT. Interestingly, we also show that the communication overheads follow FedAsync $>$ FedAT $>$ TT-Fed $>$ FedAvg, which means TT-Fed can substantially improve the communication efficiency compared to FedAsync and FedAT.
\end{itemize}

The rest of this paper is organized as follows. Section II introduces the system model. The convergence analysis of TT-Fed is given in Section III. The resource allocation problem is solved in Section IV, followed by numerical results in Section V. Section VI concludes this paper.

\section{System Model}
We consider a wireless network with an edge server and $U$ users, denoted by the set ${\cal{U}}=\{1,2,\dots,U\}$, jointly perform FL task. Each user $u \in {\cal{U}}$ possesses a local data set ${\cal{D}}_u$ with $D_u=|{\cal{D}}_u|$ data samples, and each data sample is represented by an input-output pair $(x_{u,i}, y_{u,i}), ~i \in {\cal{D}}_u$, where $x_{u,i} \in \mathbb{R}^m$ denotes the feature vector and $y_{u,i} \in \mathbb{R}$ is the corresponding ground truth.
The total number of data samples in the whole network is denoted as $D$, where $ D = \sum_{u \in {\cal{U}}} D_u$.

The next two subsections compare the differences between the event-triggered FLs (FedAvg, FedAsync, and FedAT) and our proposed TT-Fed. For clarification, we present two definitions:

\noindent \textbf{Definition 1.} (\textbf{Global Aggregation Round}) \textit{is the duration between two adjacent global aggregations.}

\noindent \textbf{Definition 2.} (\textbf{Local Updating Round}) \textit{ is duration for a user to receive global model, complete local computation and upload the model to the server.}

\subsection{Event-triggered Federated Learning}
The optimization goal of an event-triggered FL is to find a global model $w^* \in \mathbb{R}^m$ that minimizes the empirical risk
\vspace{-0.1cm}
\begin{align}
w^* = \argmin _{w \in \mathbb{R}^m} \frac{1}{D} \sum _{u \in {\cal{U}}} \sum _{i \in {\cal{D}}_u} f(w;x_{u,i}, y_{u,i}), \label{FL_global}
\end{align}
where $f(w;x_{u,i}, y_{u,i})$ is the predefined loss function on data sample $(x_{u,i}, y_{u,i})$.

The iterative learning process consists of three main steps, namely, global model transmission, local model computation, and global model aggregation:

(a) \textbf{Global Model Transmission.} At the beginning of the $j$-th global aggregation round, the server distributes current global model $w_{\text{G}}^j$ to users in the selected user set $ {\cal{S}}_{\text{sel}}$ to perform training. 

(b) \textbf{Local Computation.} After receiving the global model $w^{j}_{\text{G}}$, the selected user $u$ will begin the local model computation step based on its local data set ${\cal{D}}_u$. Then, after computation, it generates a local model\footnote{We use superscript to denote global aggregation round and subscript to denote user throughout this paper. The capital letters ``L'' and ``G'' are used to discriminate local model and global model, respectively.} $w^{k}_{\text{L},u}$ that is ready to be uploaded to the server at the $k$-th global aggregation round as 
\vspace{-0.2cm}
\begin{align}
w^k_{{\text{L}},u} = w^{j}_{\text{G}} - \lambda \mathbb{E} [ \nabla f(w^{j}_{\text{G}};x_{u,i}, y_{u,i})], \label{FL_local}
\end{align}
where $\lambda$ is the local learning rate and $w_{\text{G}}^j$ is the global model received at the $j$-th global aggregation round. In \eqref{FL_local}, $j=k-1$ for Sync FL, whereas $ j \le k-1$ for Async FL.

(c) \textbf{Global Aggregation.} Before the end of the $k$-th global aggregation round, the $u$-th user in the set $ {\cal{S}}_{\text{sel}}$  transmits its local model $w^{k}_{{\text{L}},u}$ to the server to perform global aggregation:

\begin{itemize}

\item \textbf{FedAvg} \cite{mcmahan2017communicationefficient} \textbf{(Sync FL)}:  upon receiving all the local models from users in $ {\cal{S}}_{\text{sel}}$, the server performs aggregation and generates a new global model $w^{k}_{\text{G, S}}$ using
\vspace{-0.1cm}
\begin{align}
w^k_{\text{G, S}} = \sum\limits_{u \in {\cal{S}}_{\text{sel}}} \frac{D_u w^k_{{\text{L}},u}}{\sum\nolimits_{u \in {\cal{S}}_{\text{sel}}} D_u }. \label{FL_aggr_sync}
\end{align}

\item \textbf{FedAsync} \cite{xie2020asynchronous} \textbf{(Async FL)}:  as soon as a random local model $w_{{\text{L}},u}^{k}$ is received, the server performs aggregation and generates a new global model $w^{k}_{\text{G, A}}$, using
\vspace{-0.1cm}
\begin{align}
w^k_{\text{G, A}} & =   \psi w_{{\text{L}},u}^{k} + \left( 1- \psi  \right) w^{k-1}_{\text{G, A}}, \label{FL_aggr_async}
\end{align}
where $\psi \in (0,1)$ is a mixing hyperparameter. When $\psi \ge 0.5$, the aggregation favors the new update, whereas when $\psi < 0.5$, the aggregation favors the latest global model.

\item \textbf{FedAT} \cite{Zheng2020FedAT}: In FedAT, a tiering module divides users into $M$ fixed tiers based on their processing delays. $M$ tier models corresponding to each tier are maintained at the server, as well as a global model. Local models from the $m$-th tier are aggregated to update the tier model $w_{tier_m}$ before merging to the global model, using 
\vspace{-0.1cm}
\begin{align}
w_{tier_m} = \sum_{u \in tier_m} \frac{D_u w^k_{{\text{L}},u}}{\sum\nolimits_{u \in tier_m} D_u }. \label{FL_tieraggr_fedat}
\end{align}

Then, the new model update from a random tier triggers the global aggregation asynchronously, and the server generates a new global model $w^k_{\text{G, AT}}$ using
\vspace{-0.1cm}
\begin{align}
w^k_{\text{G, AT}} =  \sum_{m=1}^M \alpha_m w_{tier_m}, \label{FL_aggr_fedat}
\end{align}
where $\alpha_m$ is a weighting factor for the $m$-th tier, $\sum_{m=1}^M \alpha_m=1$. One can easily notice the Sync intra-tier aggregation and Async inter-tier aggregation behavior of FedAT.

\end{itemize}

At last, the new global models $w^{k}_{\text{G, S}}$, $w^{k}_{\text{G, A}}$, and $w^k_{\text{G, AT}}$ will be distributed to the selected users in the $(k+1)$-th global aggregation round and the training process continues until convergence.

\subsection{Time-triggered Federated Learning (TT-Fed)}
\begin{figure}[t!]
\begin{minipage}[t]{0.45\linewidth}
    \includegraphics[scale=0.65]{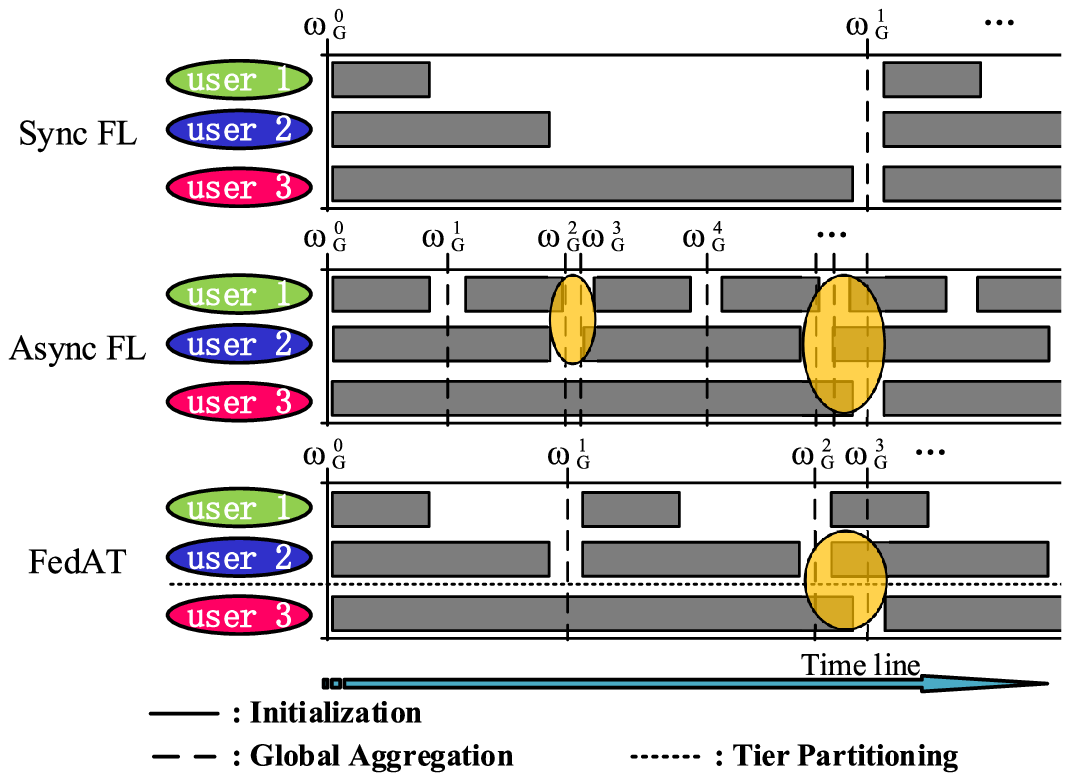}
    \caption{Comparisons of Sync FL, Async FL and FedAT. The aggregation glitch phenomenon is marked with the orange circle in Async FL and FedAT.}
    \label{Fig_intro_Async_vs_Sync}
\end{minipage}%
    \hfill%
\begin{minipage}[t]{0.45\linewidth}
    \includegraphics[scale=0.40]{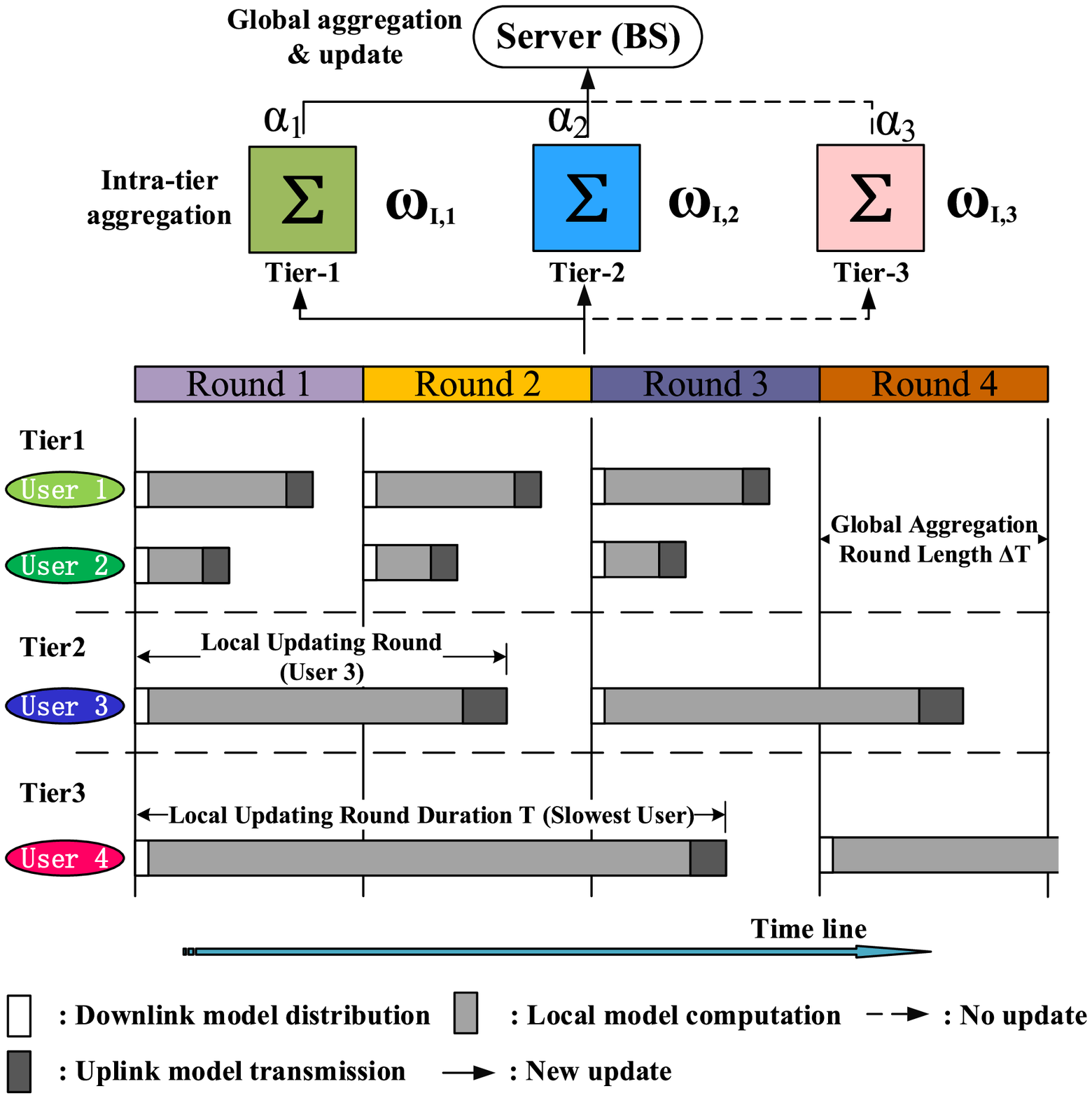}
    \caption{The work-flow of time-triggered federated learning (TT-Fed) with global aggregation round duration $\Delta T$.}
    \label{Fig_sec2_TT-Fed_Work_Flow}
\end{minipage} 
\end{figure}


The work-flow of our proposed TT-Fed is shown in Fig.~\ref{Fig_sec2_TT-Fed_Work_Flow}. Different from the event-triggered FL, the global aggregation in TT-Fed is triggered at every fixed global aggregation round duration $\Delta T$. Let us denote by $T$ the time required for the slowest user to complete one single local updating round. Then, all users are naturally partitioned into $M$ = $\lceil { \frac{T}{{\Delta T}}}\rceil$ tiers ($\lceil \cdot\rceil $ is the ceiling function), with the first tier being the fastest tier and the $M$-th tier being the slowest tier. Take an example as shown in Fig.~\ref{Fig_sec2_TT-Fed_Work_Flow}, 4 users are naturally partitioned into 3 tiers according to the global aggregation round duration partitioning. 

Let us denote by ${\cal{S}}_m$ the user set in the $m$-th tier. The tier index $m \in {\cal{M}}$ (${\cal{M}} \buildrel \Delta \over = \{1,\dots,M\}$) captures how many global aggregation rounds are needed for users in ${\cal{S}}_m$ to  complete a single local updating round. As shown in Fig. \ref{Fig_sec2_TT-Fed_Work_Flow}, user 1 and user 2 in the 1st tier ($m$=1) need a single global aggregation round ($\Delta T$) to complete their local updating, while user 4 in the 3rd tier ($m$=3) requires three global aggregation rounds ($3\Delta T$). Therefore, we have new updates from different tiers at each global aggregation round. Specifically, users in the $m$-th tier are ready to upload their local models in the $k$-th global aggregation round, if and only if $k \bmod m = 0$, where $\bmod$ denotes the modulo operation. As shown in Fig.~\ref{Fig_sec2_TT-Fed_Work_Flow}, at the end of 2nd global aggregation round ($k$=2), the server will receive local model updates from users in the 1st and 2nd tiers (i.e., $m$=1, 2), since $k \bmod m = 0$, while no update from users in the 3rd tier (i.e., $m$=3) is observed due to $k \bmod m \ne 0$.

The optimization goal of TT-Fed is the same as \eqref{FL_global}. The iterative training process contains four steps, namely, global model transmission, local computation, intra-tier aggregation, and global aggregation, which are detailed as follows:

(a) \textbf{Global Model Transmission.} At the beginning of the $(k+1)$-th global aggregation round, the server only selects users in the set $\{{\cal{S}}_m | k \bmod m $= 0,$ \, \forall m \in {\cal{M}} \}$ to perform global model transmission. This is because these users uploaded their models in the $k$-th global aggregation round and are ready for the next round.

(b) \textbf{Local Computation.} 
Since a selected user $u \in {\cal{S}}_m$ requires $m$ global aggregation rounds to complete its local updating process, the uploaded local model at the $k$-th global aggregation round is determined by the global model $w^{k-m}_{\text{G}}$. Thus, the local update of the $u$-th user is
\begin{align}
w^{k}_{\text{L},u}=w^{k-m}_{\text{G}} - \lambda \mathbb{E} [\nabla f(w^{k-m}_{\text{G}};x_{u,i},y_{u,i})], ~\forall u \in {\cal{S}}_m. \label{TT_local}
\end{align}

(c) \textbf{Intra-tier Aggregation.} Local models from the same tier are aggregated before global aggregation, and the intra-tier aggregation result for the $m$-th tier at the $k$-th global aggregation round $ w^{k}_{\text{I},m}$ is given by
\begin{align}
 w^{k}_{\text{I},m} = \sum\limits_{u \in {\cal{S}}_m} \frac{D_u w^k_{\text{L},u}}{\sum_{u \in {\cal{S}}_m}  D_u}. \label{TT_intra}
\end{align}

(d) \textbf{Global Aggregation.} 
The global aggregation at the end of the $k$-th global aggregation round is  given by
\begin{align}
w^k_{\text{G}}   =    \sum_{m=1}^M \mathds{1} \{ k\bmod m=0 \} \alpha_m^k w_{{\text{I}},m}^{k} + \sum_{m=1}^M \Big( 1-   \mathds{1} \{ k \bmod m=0 \}  \Big) \alpha_m^k  w^{k-1}_{\text{G}}, \label{TT_global}
\end{align}
where $\alpha_m^k$ is the aggregation weight of models from the $m$-th tier at the $k$-th global aggregation round, and $ \sum_{m=1}^M \alpha_m^k = 1$. In \eqref{TT_global}, $\mathds{1} \{ \cdot \} $ is the indicator function. The first term at the right hand side of \eqref{TT_global} is the weighted summation of uploaded models, and the second term is the latest global model $w^{k-1}_{\text{G}}$ multiplied by the corresponding aggregation weight. 

To balance the computation bias towards fast tiers, we adopt a heuristic weighting scheme \cite{Zheng2020FedAT}. We record the total number of updates from the $m$-th tier till the $k$-th global aggregation round, which is $\left\lfloor { \frac{k}{m}} \right\rfloor $ ($\lfloor \cdot\rfloor $ denotes the floor function), and the total updates from all tiers, which is $\sum_{m=1}^{M} \left\lfloor { \frac{k}{m}} \right\rfloor $. Intuitively, to balance the model bias, larger aggregation weights should be assigned to slower tiers, while smaller weights to faster tiers. Hence, the aggregation weight for models from the $m$-th tier at  the $k$-th global aggregation round is given by
\begin{align}
\alpha_m^k =  \frac{\big\lfloor { \frac{k}{M+1-m}} \big\rfloor}{\sum_{m=1}^{M} \big\lfloor { \frac{k}{m}} \big\rfloor }, \label{TT_weight}
\end{align}
where $\big\lfloor { \frac{k}{M+1-m}} \big\rfloor$ is the number of updates from the $(M+1-m)$-th tier. The intuition is that we swap the update times of the $m$-th tier and the $(M+1-m)$-th tier, to control the impact from the faster one and to enhance the impact from the slower one.

In our proposed TT-Fed architecture, the global aggregation, intra-tier aggregation, and local computation are connected and interrelated. Due to this cascaded connection structure and time-triggered setting, only one model needs to be stored at the server. This is fundamentally different from the event-triggered multi-tier algorithm proposed in \cite{Zheng2020FedAT}, where $(M+1)$ models need to be maintained at the server in total, with one for each tier and one for the global model.

Sync FL and Async FL are special cases of TT-Fed. To be more specific, if we set the global aggregation round length $\Delta T$ to be small enough, so that each user can be categorized into an independent tier, TT-Fed transforms to a fully Async FL. Under this circumstance, the server will update the global model whenever there is a local update. However, if we set  $\Delta T$  to be large enough (at least larger than $T$), this will result in a fully Sync setting. In such setting, users have to wait for all other users to finish local model uploading before aggregation. In this way, TT-Fed can transform into Sync and Async variations by tuning the global aggregation round duration $\Delta T$. Compared to Async FL and FedAT, TT-Fed allows users both in different tiers and in the same tier to align at the start of their local updating rounds, and thus, broadcast can be leveraged to enhance its communication efficiency. As shown in Fig. \ref{Fig_sec2_TT-Fed_Work_Flow}, at the beginning of the 3rd global aggregation round (Round 3), users in the 1st tier (user 1 and user 2) and the 2nd tier (user 3) are aligned to begin their local updating rounds. Meanwhile, as users with similar computation capabilities are grouped into smaller tiers, the straggler issues in Sync FL can be alleviated to a great extent. As shown in Fig. \ref{Fig_sec2_TT-Fed_Work_Flow}, waiting time for users in the 1st tier is largely reduced compared to that in Sync FL.


\subsection{Downlink \& Uplink Transmissions}
The downlink global model transmission takes place at the beginning of each global aggregation round via a broadcast channel. Thanks to the high transmit power of the server and the total bandwidth it can use for signal broadcasting, the time of downlink transmission is assumed to be negligible compared to that of the uplink.

The uplink local model transmission takes place before the end of each global aggregation round. We consider frequency domain multiple access (FDMA) for uplink transmission, where each selected user is allocated to a fraction of total bandwidth budget $B$ to upload their local model over different sub-channels to avoid mutual-interference. The bandwidth allocation vector at the $k$-th global aggregation round among users is denoted by $\vec{b}^{\, k} = [b_1^k, b_2^k, \dots, b_U^k]$. We consider Rayleigh fading model \cite{chen2021ajoint}, and denote $g_u^k$ to be the channel coefficient between the $u$-th user and the server during the $k$-th global aggregation round:
\begin{align} 
g_u^k= h_u^k \sqrt{l(d_u^k)}, \label{channel_gain} 
\end{align}
where $h_u^k$ represents the small-scale fading and $h_u^k \sim {\cal{CN}}(0,1)$, $d_u^k$ is the distance between the $u$-th user and the server, and $l(d_u^k)$ represents the distance-dependent path loss. In \eqref{channel_gain}, we consider non-singular path loss model $l(d) = \min(1,d^{-\alpha})$, with $\alpha \ge 2$ being the path loss factor.

Let us assume that the model size in the considered FL task is $Z$ bits. Thus, the achievable rate $r_u^k$ and the uplink communication time $\tau ^k_{u,\text{cm}}$ for the $u$-th user at the $k$-th global aggregation round are given by
\begin{align}
& r_u^k  = b_u^k \log_2 \Big(1+ \frac{P \| g_u^k \|^2  }{N_0 b_u^k} \Big),~~\forall u\in {\cal{U}}, \label{rate} \text{ and} \\
& \hspace{0.9cm} \tau ^k_{u,\text{cm}}  = \frac{Z}{r_u^k}, ~~\forall u\in {\cal{U}}, \label{delay_cm}
\end{align}
where $b_u^k$ is the bandwidth allocated to the $u$-th user at the $k$-th global aggregation round, $P$ is the uplink transmission power (we assume equal transmission power per user), and $N_0$ is the power spectral density of noise.

\subsection{Successful Transmission Probability}
We consider the successful transmission probability (STP) to characterize the uncertainty of wireless uplink transmission. We also assume that the downlink global model broadcasting is always successful, considering the wide bandwidth and high transmit power of the server. 

Let us denote by $\gamma _{\rm{th}}$ the signal to noise ratio (SNR) threshold for successful data decoding. Denote the transmission success indication variable for the selected user $u$ at the $k$-th global aggregation round to be $\rho_u^k \buildrel \Delta \over = \mathds{1}(\text{SNR}^k_u \ge \gamma _{\rm{th}})$, where $\rho_u^k =1$ means the uploaded local model can be successfully decoded at the server, while $\rho_u^k =0$ otherwise. Thus, the STP for the $u$-th user at the $k$-th global aggregation round is \cite{choudhury2007outage}
\begin{align}
& \mathbb{P}[\text{SNR}_u^k \ge \gamma _{\rm{th}}]  = \mathbb{P} \Big[ \frac{\| g_u^k \|^2 P}{N_0 b_u^k} \ge \gamma _{\rm{th}} \Big] \nonumber \\
& \hspace{0.2cm} \mathop = \limits^{(a)} \mathbb{P} \Big[ \| h_u^k \|^2 \ge \frac{\gamma _{\rm{th}} N_0 b_u^k }{P l(d_u^k)} \Big] \mathop = \limits^{(b)} e^{-\frac{\gamma _{\rm{th}} N_0 b_u^k }{P l(d_u^k)}}, \label{STP}
\end{align}
where $(a)$ is from plugging into \eqref{channel_gain}, and $(b)$ is from the cumulative probability of exponential distributed random variable $\| h_u^k \|^2$ with mean 1. We note here that the local model of user $u$ can be successfully aggregated only if this user is selected and $\rho_u^k =1$ at the same time.

\subsection{Computation Model}
We assume the CPU frequency of the $u$-th user to be $f_u$ and denote by $c_u$ the number of CPU cycles required to process a single data sample. Hence, the computation time for the $u$-th user to complete its local update in the $k$-th global aggregation round is given by \cite{yang2021energy}
\begin{align}
\tau ^k_{u,\text{cp}} = \zeta \frac{D_u c_u}{f_u}, 
\end{align}
where $\zeta$ is the number of local training epochs. We limit the computation time to be mainly local computations, whereas the computation time at the server is ignored, considering the abundant computation resource at the server and the low complexity of global aggregation.

\subsection{Problem Formulation}
An optimization problem is formulated to jointly perform  bandwidth allocation and  user selection, with the aim of minimizing the global loss function as
\begin{subequations}\label{P1_goal}
\begin{align}
\text{(P1)} \quad \min\limits_{\mathbf{a},\mathbf{b}}\; &  \frac{1}{D} \sum _{u \in {\cal{U}}} \sum _{i \in {\cal{D}}_u} f(w;x_{u,i}, y_{u,i})   \tag{\ref{P1_goal}} \\
s.t.\; &  \vec{a}^{\, k} \in \{0,1\} ^U , \; \forall k \in {\cal{K}} ,  \label{P1_1}\\
& 0 \le {b_u^k} \le  a_u^k B, \; \forall k\in {\cal{K}}, \forall u \in {\cal{U}}, \label{P1_2} \\
& \sum\nolimits_{u\in {\cal{U}}} b_u^k \leq B,\; \forall k \in {\cal{K}}, \label{P1_3} \\
& a_u^k \left( \tau ^k_{u,\text{cm}} + \tau ^k_{u,\text{cp}} \right) \le m \Delta T, \forall u \in {\cal{S}}_m, \forall m \in {\cal{M}}, \label{P1_4}
\end{align}
\end{subequations}
where $\mathbf{a} {\buildrel \Delta \over =}  [\vec{a}^{\, 1}, \vec{a}^{\, 2},...,\vec{a}^{\, K}]$ and $\mathbf{b} {\buildrel \Delta \over =} [\vec{b}^{\, 1}, \vec{b}^{\, 2}, \dots, \vec{b}^{\, K}]$ are the user selection and bandwidth allocation strategy for the whole training process, and ${\cal{K}} \buildrel \Delta \over = \{1, \dots, K \}$ where $K$ refers to the number of global aggregation rounds for TT-Fed to converge. The minimization goal \eqref{P1_goal} is the global training loss. \eqref{P1_1} is the inherent constraint for user selection vector at the $k$-th round, $\vec{a}^{\, k} {\buildrel \Delta \over =}  [{a}^k_1, {a}^k_2,...,{a}^k_U]$, where $a^k_u = 1$ means user $u$ is selected to update in the $k$-th round, and $a^k_u =0$ otherwise. \eqref{P1_2} ensures that no bandwidth is wasted to an unselected user. \eqref{P1_3} means the total allocated bandwidth at each global aggregation round is within the bandwidth budget $B$. \eqref{P1_4} ensures that a selected user can upload its local model before the aggregation deadline, where $m \Delta T$ stands for $m$ training rounds required in total for user $u \in {\cal{S}}_m$ to complete a single local updating round.

Problem (P1) is a mixed integer programming problem. To solve it, we need to find an analytical expression of the optimization goal with respect to all the optimization variables. However, since the wireless channel and potential update tiers vary over different global aggregation rounds, it is impossible to find a tractable expression, making it impractical to solve (P1) directly.

\section{Convergence Analysis}
To solve (P1), we start from the convergence analysis of TT-Fed to analyze how wireless communication affects its convergence speed and accuracy. To avoid tedious derivation and grab the main points, we assume that the user selection for each tier, and the bandwidth allocation for each selected user remains unchanged during the whole training process. We will revisit the user selection and bandwidth allocation problem in section IV, without the above assumptions. 

For notation simplicity, we denote $F(w)$ to be the global training loss as shown in \eqref{P1_goal}. Denote by ${\cal{S}}_{m,\text{S}} \buildrel \Delta \over = \{u|a_u^k \rho_u^k=1, \forall u \in {\cal{S}}_m \}$ the users in the $m$-th tier who have successfully uploaded their parameters and ${\cal{S}}_{m,\text{F}}  \buildrel \Delta \over = {\cal{S}}_m \backslash {\cal{S}}_{m,\text{S}}$ the rest of the users who have failed in uploading their models in the same tier. We denote the total number of data samples in sets ${\cal{S}}_m,~{\cal{S}}_{m,\text{S}},{\cal{S}}_{m,\text{F}}$ to be $D_m$ = $\sum_{u \in {\cal{S}}_m} D_u$, $D_{m,\text{S}}$ = $\sum_{u \in {\cal{S}}_m} a_u^k \rho_u^k D_u$, and $D_{m,\text{F}}$ = $D_m -D_{m,\text{S}}$, respectively.

Before the convergence analysis, we make the following assumptions


\begin{itemize}
\item 
The global loss function $F$ is $L$-smooth, i.e. $\forall x,y$:
\vspace{-0.2cm}
\begin{align}
F(y)-F(x) \leq (y-x)^\intercal \nabla F(x) + \frac{L}{2} \| y-x \|^2. \label{assum:smooth}
 \end{align}
 The smoothness assumption provides an assurance that the gradient does not change too quickly, so that the old gradient can provide us with useful information about a nearby new gradient if we take a small step.

\item 
 $F$ is $ \mu$-strong convex, i.e. $\forall x,y$:
\vspace{-0.2cm}
\begin{align}
F(y)-F(x) \geq (y-x)^\intercal \nabla F(x) + \frac{\mu}{2} \| y-x \|^2. \label{assum:cvx}
\end{align}
The convexity assumption allows us to transform the gradient information into distance-related information, which will make the mathematical derivation process more tractable.

\item  Bounded local and global gradient dissimilarity, i.e.:
\vspace{-0.2cm}
\begin{align}
 \|\nabla f(w^{k}_{\text{G}};x_{u,i},y_{u,i})\|^2  \le \chi + \nu \|\nabla F(w^{k}_{\text{G}})\|^2,  \label{assum:local_global}
\end{align} 
where $\chi, \nu$ are positive constants. The above three assumptions are widely used in the convergence analysis for Sync FL, where the first two have been used in \cite{Li2020OnTC,yang2021energy} and the third assumption is used in \cite{chen2021ajoint} to characterize the dissimilarity of the local and global training loss at the same global aggregation round.

\item  Bounded global gradient change within $m \in {\cal{M}}$ training rounds, i.e.:
\vspace{-0.1cm}
\begin{align}
 (w^{k-m}_{\text{G}} - w^{k-1}_{\text{G}}) ^\intercal \nabla F(w^{k-1}_{\text{G}}) &\le \delta \| \nabla F(w^{k-1}_{\text{G}}) \|^2, \label{assum:global_1} 
 \end{align}
 \begin{align}
 \| w^{k-m}_{\text{G}} - w^{k-1}_{\text{G}} \| &\le \varepsilon, \label{assum:global_2}
\end{align}
where $\delta$ and $\varepsilon$ are positive constants. \eqref{assum:global_1} can easily be satisfied based on \eqref{assum:global_2} and Cauchy-Schwartz inequality, since the left hand side of the inequality is the inner product of directions $(w^{k-m}_{\text{G}} - w^{k-1}_{\text{G}})$ and $\nabla F(w^{k-1}_{\text{G}})$.

\item  Bounded local gradient within $m \in {\cal{M}}$ training rounds, i.e.:
\vspace{-0.1cm}
\begin{align}
\| \nabla f(w^{k-m}_{\text{G}}) \| \le \beta \| \nabla f(w^{k-1}_{\text{G}}) \|, \label{assum:local_1} \\
\|\nabla f(w^{k-1}_{\text{G}}) - \nabla f(w^{k-m}_{\text{G}}) \| \le \phi , \label{assum:local_2}
\end{align}
where $\beta$ and $\phi$ are both positive constants. The above two assumptions ensure that the model changes within $M$ training rounds are bounded. Similar assumptions are commonly seen in the convergence analysis of Async FL \cite{xie2020asynchronous,Zheng2020FedAT}. When $M=1$, TT-Fed transforms to Sync FL with only one tier. In this scenario, equations \eqref{assum:global_1}-\eqref{assum:local_2} are automatically satisfied with $\beta = 1$, $\delta = \varepsilon=\phi=0$ and can be left out, meaning that the above assumptions are compatible to Sync settings.
\end{itemize}

The above assumptions can be readily satisfied by widely used loss functions. Moreover, although we assume $F$ to be convex, our simulation results show TT-Fed works well under non-convex loss functions, e.g., cross-entropy of the neural network. The training loss of TT-Fed is upper bounded by the following theorem.

\begin{theo}
Under fixed user selection and bandwidth allocation strategy for TT-Fed, the performance gap between the loss function $F(w^K_{\text{G}})$ at the $K$-th global aggregation round and the optimum value $F(w^*_{\text{G}})$ is upper bounded by
\vspace{-0.1cm}
\begin{align}
\mathbb{E} \left\{ F(w^K_{\text{G}})-F(w^*_{\text{G}}) \right\} \le \mathbb{E} \bigg \{ \underbrace {\big( 1-\frac{\mu\xi}{2L}\Delta_2 \big)^K  \left[F(w^{0}_{\text{G}}) - F(w^*_{\text{G}})\right]}_{\text{first term}} + \underbrace { 2\frac{\Delta_1 L}{\mu \Delta_2} \Big[1-\big( 1- \frac{\mu\xi}{2L}\Delta_2 \big)^K \Big] }_{\text{second term}} \bigg\} , \label{TT_bound}
\end{align}
where 
\vspace{-0.2cm}
\begin{align}
\Delta_1 & = \frac{1}{M}\sum\nolimits_{m=1}^{M} L\varepsilon^2 + \frac{3}{4L}\Big[ \phi^2+ \chi \Big( 1+ \frac{(1+\beta)^2 D_{m,\text{F}}}{D_m} \Big) \Big]  ,  \label{delta1}\\
\Delta_2 & = \frac{1}{M}\sum\nolimits_{m=1}^{M} 1-4\delta L -3\nu\Big[ 1+ \frac{(1+\beta)^2 D_{m,\text{F}}}{D_m} \Big], \label{delta2}
\end{align}
$\xi$ is a constant and $\xi \in (0,M)$. 
\label{TT_convergence}
\end{theo}
\vspace{-0.3cm}
\begin{proof}
See Appendix A.
\end{proof}

The convergence of TT-Fed can be assured if the term $( 1-\frac{\mu\xi}{2L}\Delta_2 )$ is within range $(0,1)$. Thus, we have the following proposition to guarantee the convergence of TT-Fed.
\begin{ppro}
The proposed time-triggered FL is guaranteed to converge if the following conditions are satisfied:
\vspace{-0.2cm}
\begin{align}
& \hspace{0.9cm} 0 \le \frac{\mu}{2L} \le \frac{1}{M}, \label{cor1_1} \text{ and}\\
& 0 \le 4\delta L +3 \nu [1+(1+\beta)^2] \le 1. \label{cor1_2}
\end{align}
\label{convergence_con}
\end{ppro}
\vspace{-1cm}
\begin{proof}
From \textbf{Theorem \ref{TT_convergence}}, we can see that if $0 \le ( 1-\frac{\mu\xi}{2L}\Delta_2 ) \le 1$, then
\begin{align}
\lim_{K \to \infty} \big( 1-\frac{\mu\xi}{2L}\Delta_2 \big)^K = 0. \label{rate_limit}
\end{align}

 This ensures that the convergence gap decreases as the number of training rounds increases. Given that the total global aggregation round $K$ is large enough, the first term in the upper bound \eqref{TT_bound} will diminish and the algorithm eventually reaches a stationary point, as specified by the second term (which is a constant). To guarantee $0 \le \frac{\mu\xi}{2L}\Delta_2 \le 1$, we need to ensure $0 \le \frac{\mu\xi}{2L} \le  1$, and $0 \le \Delta_2 \le  1$. Noticing $\frac{\mu\xi}{2L} \le \frac{\mu M}{2L}$, and $\frac{D_{m,\text{F}}}{D_m} \le 1$ in \eqref{delta2}, the results can be obtained.
\end{proof}

We note that \textbf{Proposition \ref{convergence_con}} offers a sufficient instead of necessary condition. In other words, there are feasible hyperparameter settings that guarantee the convergence of TT-Fed, but $\frac{\mu\xi}{2L}\Delta_2 $ does not fall in the range of $ (0,1)$. 

Important insights can be gained from \textbf{Theorem \ref{TT_convergence}} on how wireless communication affects the convergence rate and accuracy of TT-Fed. As it can be seen from \textbf{Proposition \ref{convergence_con}}, the smaller the term $( 1-\frac{\mu\xi}{2L}\Delta_2 ) \in (0,1)$, the faster the algorithm converges. To minimize this term means to maximize $\Delta_2$, due to $\frac{\mu\xi}{2L} \ge 0$. According to \eqref{delta2}, this requires us to minimize the data sample amount of users who failed in uploading their models $D_{m,\text{F}}$. Thus, $D_{m,\text{F}}$ reveals the impact of local updates uplink transmission on the convergence rate. Moreover, we know from \textbf{Proposition \ref{convergence_con}} that the second term at the right hand side of \eqref{TT_bound} affects the convergence accuracy. As $K \to \infty$, we have \eqref{rate_limit} hold. Thus, to enhance the convergence accuracy, the term $\frac{\Delta_1 L}{\mu \Delta_2}$ should be minimized. This can be done by minimizing $\Delta_1$ while maximizing $\Delta_2$. According to \eqref{delta1} and \eqref{delta2},  this also requires us to minimize $D_{m,\text{F}}$. Therefore, minimizing the number of unsuccessfully updated users is both helpful in boosting convergence rate and convergence accuracy. This obtained result is in line with the conclusions of existing works \cite{shi2021joint,howard2020scheduling,Nishio2019client}.

\noindent \textbf{Remark 1. }
The obtained convergence analysis in \textbf{Theorem \ref{TT_convergence}} reveals how the number of tiers $M$ affects the convergence accuracy and the time needed for the algorithm to converge. As mentioned above, the term $\big( 1-\frac{\mu\xi}{2L}\Delta_2 \big)$ influences the convergence rate, while the term $\frac{\Delta_1 L}{\mu \Delta_2}$ influences the convergence accuracy. To mitigate the negative impact brought by failed uploading of local models, we assume all users can successfully upload their local models, i.e., $D_{m,\text{F}}=0$. 

For the special case of $M=1$, we have only one tier in the system, which is equivalent to Sync FL, hence $\phi =\varepsilon =\delta=0$. According to \eqref{delta1} and \eqref{delta2}, $\Delta_1$ reaches its minimum and  $\Delta_2$ reaches its maximum at the same time. Thus, the convergence accuracy can be maximized when $K \to \infty$. This is as expected, because the models from all users are synchronized in each global aggregation round and the global model can gain the full-scale information from all users in each round. However, the global aggregation round interval $\Delta T$ need to be large enough to include all users into the same tier, which is undesirable, since the existence of stragglers may prolong the whole convergence time. 

For the special case of $M = U$, we have only one user in each tier, which becomes Async FL. Under this scenario, the parameters $\varepsilon, \delta$ and $\phi$ are forced to deviate a lot from $0$, due to the difference in computation capability for each user. In other words, the local update from a slow user could be based on a fairly old global model, which may lead to a totally different or even opposite optimization direction to the current global model. Moreover, the non-IID data nature of FL intensifies this kind of deviation, and leads to degraded convergence accuracy. However, since the training interval $\Delta T$ is very small, the straggler issue can be effectively solved. 

In summary, if we carefully tune our global aggregation round partitioning, so that $\varepsilon, \delta$ and $\phi$ do not deviate much from $0$ and $\Delta T$ is much smaller than $T$, it is possible to reap the gains in terms of both convergence accuracy and convergence time.

\section{Resource Allocation for TT-Fed}
In this section, we aim to decompose the intractable optimization problem (P1) into tractable sub-problems. We consider hyperparameter settings, which satisfy \textbf{Proposition \ref{convergence_con}} to ensure the convergence of TT-Fed.

Suppose the training of TT-Fed terminates at training round $K$. Then, according to our convergence analysis in Section III, minimizing the optimization goal (16) of (P1) is equivalent to minimizing the convergence upper bound (24). Based on our analysis and \textbf{Remark 1} in Section III, to minimize the convergence upper bound (24) and increase the convergence speed of TT-Fed, we need to increase the number of successfully transmitted users. Therefore, taking the aggregation weights and local data set sizes of different users into account, problem (P1) can be decomposed to sub-problem ($\text{P}_{\text{k}}$) for global aggregation round $k$ as

\vspace{-0.5cm}
\begin{subequations}\label{Pg_goal}
\begin{align}
(\text{P}_{\text{k}}) \; \max\limits_{\vec{b}^{\, k},\vec{a}^{\, k}}\, &  \sum\nolimits _{m=1}^M \sum\nolimits _{u \in {\cal{S}}_m} \mathds{1}\{k \bmod m =0\} \alpha_m^k a^k_u D_u e^{-\frac{\gamma_{\rm{th}} N_0 b_u^k }{P l(d_u^k) }}    \tag{\ref{Pg_goal}} \\
s.t.\, &  0 \le a_u^k \le \mathds{1}\{k \bmod m=0\}, \, \forall u \in {\cal{S}}_m, m \in {\cal{M}}, \label{Pg_1}\\
& a_u^k \in \{0,1\} , \, \forall  u \in {\cal{U}},  \label{Pg_2} \\
& 0 \le \nicefrac{b_u^k}{B} \le a_u^k, \, \forall  u \in {\cal{U}}, \label{Pg_3} \\
& \sum\nolimits_{u=1}^U b_u^k \leq B, \label{Pg_4}\\
& a_u^k \left( \tau ^k_{u,\text{cm}} + \tau ^k_{u,\text{cp}} \right) \le m \Delta T, \, \forall u \in {\cal{S}}_m, ~ m \in {\cal{M}}, \label{Pg_5}
\end{align}
\end{subequations}
where the indicator function $\mathds{1}\{k\bmod m =0\}$ in the optimization goal means the model from the $m$-th tier is uploaded only at training rounds with $\{k \bmod m =0,~\forall k \in {\cal{K}} \}$. The optimization goal \eqref{Pg_goal} is the expectation value of the total data samples of successfully updated users, i.e. $\sum_{m }^M D_{m,\text{S}}$, weighted by $\alpha_m^k$. The constraint \eqref{Pg_1} ensures that users who are not supposed to upload at the $k$-th global aggregation round are not selected, which avoids potential resource waste. The rest of the constraints are the same as those in problem (P1).

Clearly, although the optimization goal in ($\text{P}_{\text{k}}$) is simplified compared to problem (P1) and is decomposed with respect to a single global aggregation round, problem ($\text{P}_{\text{k}}$) is still a mixed integer programming problem. To solve it, we decouple this problem into user selection and resource allocation sub-problems respectively, and then propose an online search algorithm to solve the joint user selection and resource allocation problem in the following \textbf{Algorithm 1}.

\begin{algorithm}
\caption{Online User Selection Algorithm based on Optimum Bandwidth Allocation }\label{online}
\begin{algorithmic}[1]
\State Initialize user selection vector $\vec{a}^{\, k}$ and bandwidth allocation vector $\vec{b}^{\, k}$ to $\mathbf{0}$; Initialize allocated bandwidth amount $B_{\text{allo}} = 0$; Initialize selected user set ${\cal{S}}_{\text{sel}}^k = \emptyset $ and qualified user set ${\cal{S}}_{\text{qual}}^k = \emptyset $ for current global aggregation round
\For   {$ m \in \{1,\dots,M\}$}
        \If {$k \bmod m$ = 0}
            \For {$\forall u \in {\cal{S}}_m$}
                \State ${\cal{S}}_{\text{qual}}^k = {\cal{S}}_{\text{qual}}^k  \cup \{u \} $
                \State Calculate $\Lambda_u^k $ using \eqref{Lambda}; Initialize: $\tilde b^{k}_u$ using \eqref{opt_bandwidth}
                \State Initialize $ \textit{Weight}_u^k = \alpha_m^k D_u e^{-\frac{\gamma_{\rm{th}} N_0 \tilde b_u^k }{P l(d_u^k)}} $
            \EndFor
        \EndIf
\EndFor

\While {True}
        \State  $u^* = \argmax _{u \in {\cal{S}}_{\text{qual}}^k} \textit{Weight}_u^k$ 
        \If {$B_{\text{allo}} + \tilde b^{k}_{u^*} \le B$}
                \State Update $b^{k}_{u^*} = \tilde b^{k}_{u^*}$, $a^k_{u^*} = 1$, $B_{\text{allo}} = B_{\text{allo}} + b^{k}_{u^*}$; Update ${\cal{S}}_{\text{qual}}^k = {\cal{S}}_{\text{qual}}^k  \backslash \{ {u^*} \}$
        \Else 
                \State Break
        \EndIf

        \If {${\cal{S}}_{\text{qual}}^k = \emptyset$}
            \State Break
        \EndIf
\EndWhile
\Return Vectors $\vec{a}^{\, k}$, $\vec{b}^{\, k}$.
\end{algorithmic}
\end{algorithm}

We first optimize the bandwidth allocation under the fixed user selection vector $\vec{a}^{\, k}$. 
\vspace{-0.1cm}
\begin{theo}
Under the given user selection strategy $\vec{a}^{\, k}$ with unlimited bandwidth, the optimum bandwidth allocation for selected user $u$  is given by
\vspace{-0.1cm}
\begin{align}
(b^{k}_u)^* = \frac{-Z \ln 2 }{\left[ W_{-1}\left( -\Lambda_u^k e^{-\Lambda_u^k}\right) + \Lambda_u^k \right] \left( m \Delta T - \tau_{u,\text{cp}}^k \right)},     \label{opt_bandwidth}
\end{align}
where
\vspace{-0.1cm}
\begin{align}
\Lambda_u^k = \frac{Z N_0 \ln 2}{P \| g_u^k \|^2 \left(m \Delta T - \tau _{u,\text{cp}}^k \right)}, \label{Lambda}
\end{align}
$W_{-1}(\cdot)$ denotes the $W_{-1}$ branch of the Lambert-$W$ function \cite{Corless96onthe}.
\label{bandwidth_allo}
\end{theo}
\vspace{-0.3cm}
\begin{proof}
See Appendix B.
\end{proof}

Due to the scarcity of wireless network resources, if the total amount of allocated bandwidth for the selected users exceeds the budget, we need to exclude less important users in maximizing the optimization goal. On the other hand, given enough unallocated bandwidth, it is possible to include additional potential users to make the best use of resources. Hence, we propose an online user selection algorithm as \textbf{Algorithm 1} based on optimum bandwidth allocation to solve problem ($\text{P}_{\text{k}}$). In steps 2 - 7 of \textbf{Algorithm 1}, we initialize the parameter settings for users in the qualified user set ${\cal{S}}_{\text{qual}}^k$ that managed to complete their local updatings at the $k$-th global aggregation round. In step 6, the optimum bandwidth allocation for each qualified user under unlimited bandwidth budget is obtained based on \textbf{Theorem \ref{bandwidth_allo}}. Then, we calculate the contribution weight of each qualified user in maximizing the goal of ($\text{P}_{\text{k}}$). In steps 8 - 15, We iteratively select the qualified user with highest contribution weight until the total allocated bandwidth $B_{\text{allo}}$ reaches the bandwidth budget $B$ or all qualified users are selected.

The time complexity of the proposed resource allocation algorithm consists of two parts: 1) the optimum bandwidth allocation given by Theorem 2, whose time complexity is $\mathcal{O}(U)$; and 2) the online user selection algorithm given by Algorithm 1, whose worst case time complexity is $\mathcal{O}(U^2)$, where each user forms an independent tier. Therefore, the total time complexity at the server side is in the order of $\mathcal{O}(U^2)$, and the space complexity is in the order of $\mathcal{O}(U)$.

\section{Numerical Results}
In the simulations, we consider a cellular network with coverage radius $R=600 \rm{m}$, inside which 20 users are randomly and uniformly distributed. 
The noise power spectrum density is $N_0 = -174\rm{dBm/Hz}$ and we set the path loss factor as $\alpha = 3.76$ \cite{shi2021joint}. The SNR threshold for successful decoding is $\gamma _{\rm{th}}=0 \rm{dB}$. The transmit power of each user is set to be $P = 10\rm{mW}$ and the total bandwidth budget for uplink wireless transmission is $B = 20\rm{MHz}$. 
Each user trains a feedforward neural network (FNN) with a single hidden layer that contains 50 neurons using the cross-entropy loss function. We train the FNNs using different FL algorithms via a subset of the MNIST database \cite{lecun1998mnist}, which contains $2500$ equal sized 10 classes of 0-9 hand-written digit training images. The learned FL models are then evaluated on the test set of MNIST, which contains $10,000$ test images.  Unless otherwise specified, the local CPU frequency of each user is $1\rm{GHz}$ and it requires $5 \times 10^5$ CPU cycles to process one data sample. 
We use 16 bits to represent each model parameter. For performance comparisons, we consider three state-of-the-art event-triggered FL algorithms as the benchmarks, namely, FedAvg in \cite{mcmahan2017communicationefficient}, FedAsync in \cite{xie2020asynchronous} and FedAT in \cite{Zheng2020FedAT}. Their detailed training processes have been provided in Section II-A.

\begin{figure}[t!]
\begin{minipage}[t]{0.45\linewidth}
    \includegraphics[scale=0.55]{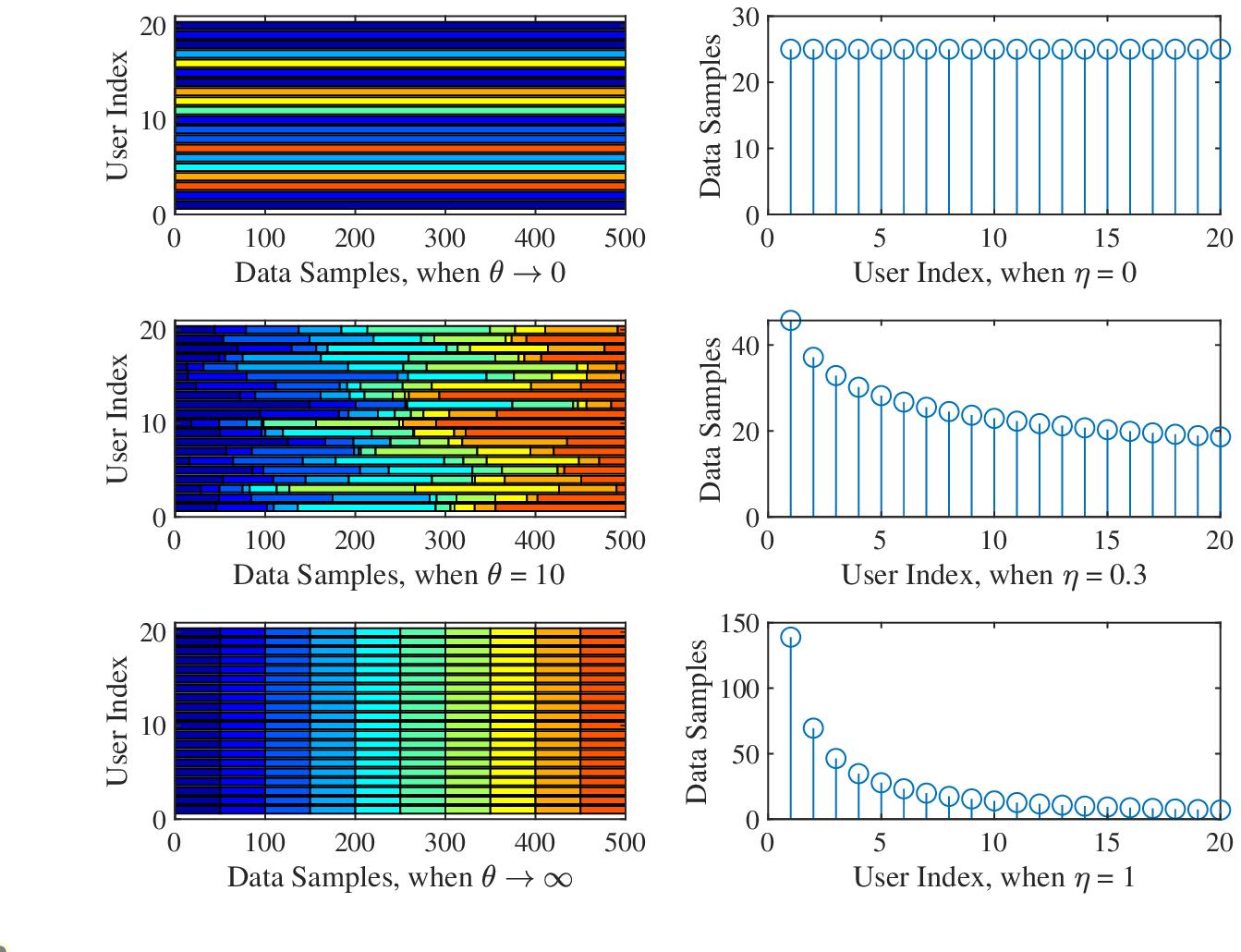}
    \caption{An illustration of applied non-IID modeling.}
    \label{Fig_sec5_non_iid_illustration}
\end{minipage}%
    \hfill%
\begin{minipage}[t]{0.45\linewidth}
    \includegraphics[scale=0.55]{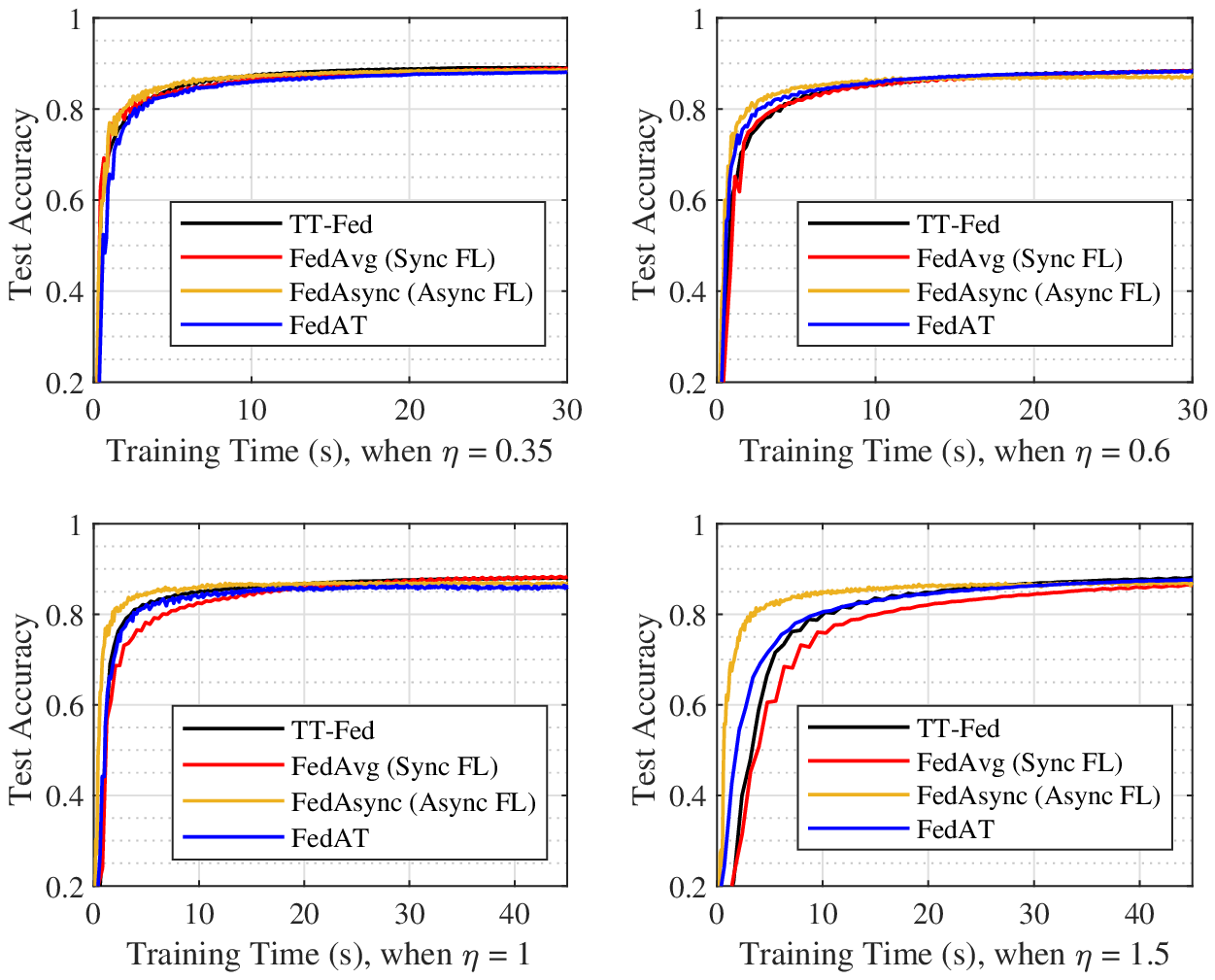}
    \caption{Test accuracy under different Zipf parameters.}
    \label{Fig_sec5_Zipf_accuracy_comparisons}
\end{minipage} 
\end{figure}

To characterize non-IID data behavior of users, we adopt two data distribution models \cite{wadu2021joint}:

\emph{1) Heterogeneous data classes.} We use the Dirichlet distribution $q_n = \frac{v_n}{\sum_{n=1}^N v_n}$ to characterize the varying data classes owned by different users \cite{yurochkin2019bayesian,hsu2019measuring}, where $q_n \in [0,1]$ is a random variable denoting the share of the $n$-th data class, $N$ is the number of total data classes, $v_n$ is a Gamma-distributed random variable drawn from $\mathrm{Gamma}(\theta \bar q_n, 1)$ with $\theta$ being the concentration parameter and $\bar q_n$ being the original distribution of class $n$ in the whole system data set, i.e., $\bar q_n =0.1$ for our constructed training set. The effect of different concentration parameter setting is shown in the left column of Fig. \ref{Fig_sec5_non_iid_illustration}. We use different color bars to represent different data classes. It is observed that when $\theta  \to 0$, each user will randomly possess only one class of data samples; while when $\theta \to \infty$, each user will possess equal sized data samples for each class.

\emph{2) Unbalanced data amounts.} We use the Zipf distribution $D_u =  \frac{D u^{-\eta}}{\sum_{u \in {\cal{U}}} u^{-\eta}}$ to characterize the different number of data samples owned by different users \cite{powers1998applications}, where the parameter $ \eta $ is used to characterize the skewness of the data set size for different user. The effect of different Zipf parameter setting is shown in the right column of Fig. \ref{Fig_sec5_non_iid_illustration}. When $ \eta = 0 $, each user will possess a data set with equal size $ \frac{D}{U} $. Continue increasing $\eta$ will make user 1 possess most of the data samples in the network. When $ \eta \to \infty $, the local data set of user 1 contains all the data samples, whereas other users possess no data.

\subsection{Impact of data distribution}
This subsection analyzes the impact of data distribution on the test accuracies of all considered FL algorithms. We set $\Delta T=0.6T$ for TT-Fed, so that the 20 users are naturally grouped into 2 tiers. For fair comparison, each tier in FedAT contains the same users as that in TT-Fed.

Fig. \ref{Fig_sec5_Zipf_accuracy_comparisons} shows the impact of unbalanced data amounts. The test accuracy comparisons of different FL algorithms are conducted under different Zipf parameter $\eta$, whereas the concentration parameter is fixed $\theta \to \infty$, which means each user will possess different size of local data set with identical data class distribution. It is observed that, as the training time increases, the test accuracies of TT-Fed and the considered benchmarks increase and eventually converge to stable levels. Moreover, we notice that FedAsync converges quicker than all the other three algorithms under various Zipf parameters $\eta$. This is because their global aggregation frequencies  follow  FedAsync $>$ FedAT $>$ TT-Fed $>$ FedAvg. When $\eta$ is small, i.e., $\eta=0.35,~0.6$, the convergence speeds and accuracies of these four algorithms are similar. This is because the computation delays among users are similar, and thus having more tiers is not beneficial. As $\eta$ increases, i.e., $\eta=1,~1.5$, the convergence speeds of FedAT, TT-Fed and FedAvg slow down, which is due to the intensified straggler issues. In summary, under highly skewed data set size setting with identical data class distribution, since the data set of a random user contains the full information of the global data class distribution, FL with more tiers obtains faster convergence speed.

\begin{figure}[t!]
\begin{minipage}[t]{0.45\linewidth}
    \includegraphics[scale=0.55]{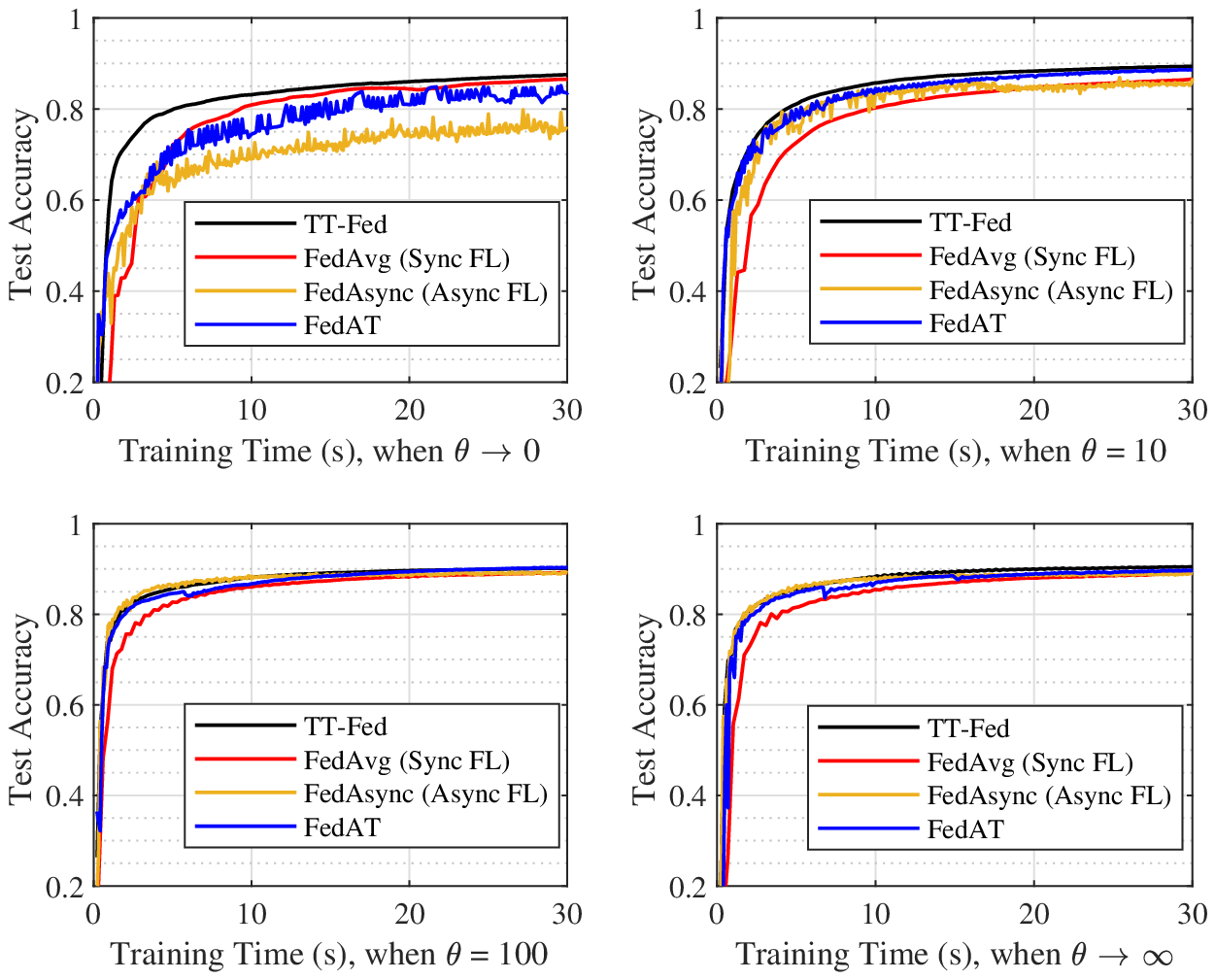}
    \caption{Test accuracy under different concentration parameters.}
    \label{Fig_sec5_Dirichlet_accuracy_comparisons}
\end{minipage}%
    \hfill%
\begin{minipage}[t]{0.45\linewidth}
    \includegraphics[scale=0.55]{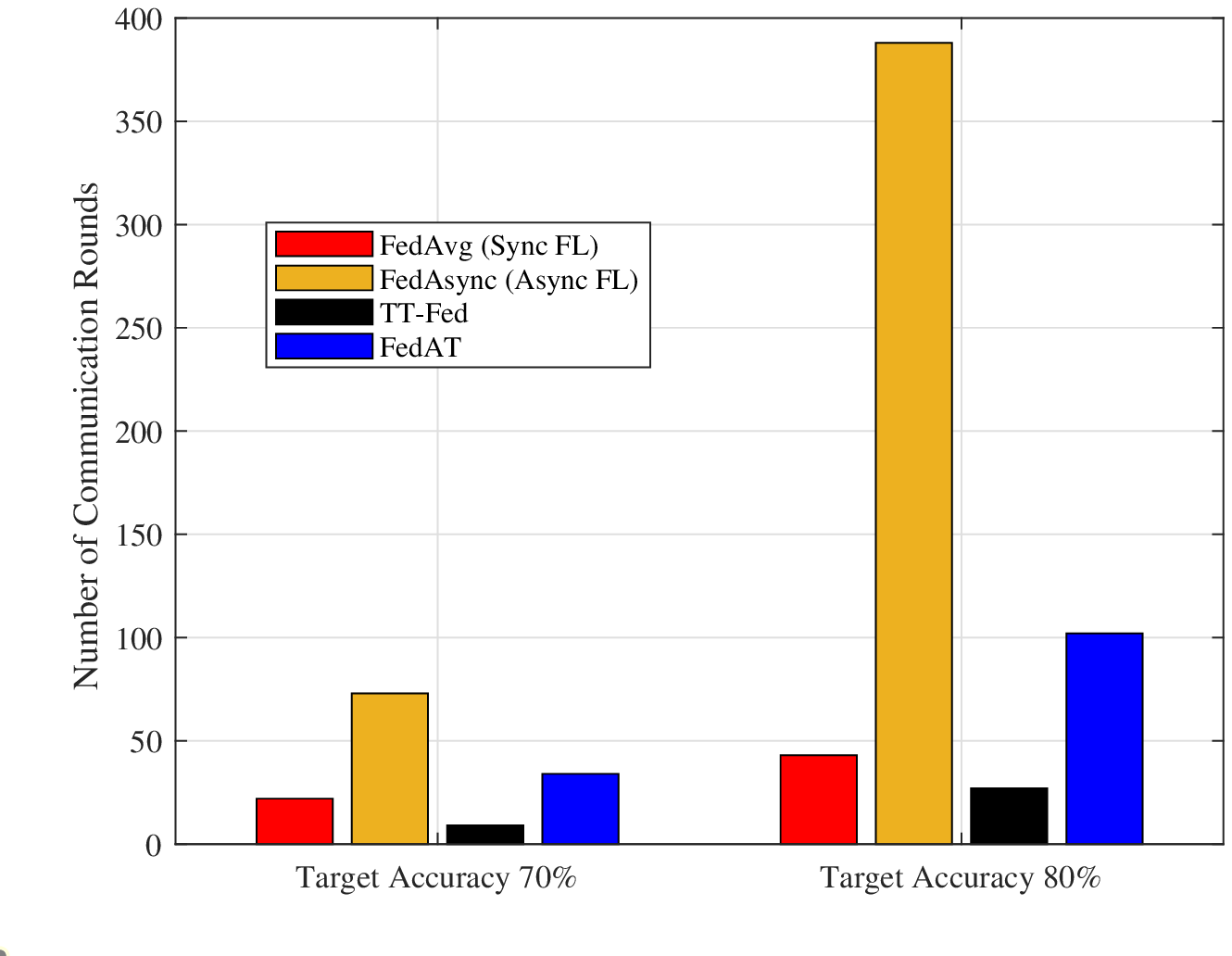}
    \caption{Communication rounds required to achieve a given target accuracy.}
    \label{Fig_sec5_Communication_Round_Comparisons}
\end{minipage} 
\end{figure}

Fig. \ref{Fig_sec5_Dirichlet_accuracy_comparisons} plots the impact of  heterogeneous data classes. We compare the test accuracy of considered FL algorithms under various concentration parameter $\theta$, with $\eta = 0$, which means each user has equal-sized local data set with skewed data class distribution specified by $\theta$. As each user possesses an equal-sized local data set, to characterize the difference in computation delay, the CPU frequency of each user is randomly drawn from the range of $1 \sim 5 ~ \rm{GHz}$. 

When $\theta$ is large, i.e., $\theta \to \infty,~ \theta=100$, each user possesses nearly evenly distributed data classes. In this case, FedAsync obtains the highest convergence speed. As $\theta$ decreases, i.e., $\theta \to 0,~ \theta=10$, the convergence accuracies of all algorithms degrade with prolonged convergence time, which is due to the fact that under highly non-IID data the server needs more communication rounds to converge.  Moreover, the learning curves of FedAsync and FedAT fluctuate. This is because the data class distribution at each user becomes more and more skewed, which means some data classes are too scarce or even missing. To have a smooth and robust learning curve, we need more information from more users for a single global aggregation step. As aggregation in FedAsync and FedAT involves fewer users and due to the existence of aggregation glitch in the event-triggered setting, updates from faster users/tiers tend to drift from the global optimum making the models not robust to skewed data. Without an efficient way to control the computation bias, the test accuracy of FedAsync is largely degraded under the highly skewed data setting.

Interestingly, even when the data class distribution becomes skewed, TT-Fed can still maintain fast convergence with robust performance, and achieve the highest accuracy. This is because the cross-tier synchronization and user grouping in TT-Fed can help gain more diverse information in a single global aggregation round and help yield a smoother learning curve compared to the event-triggered FedAT. Meanwhile, grouping users into smaller tiers can efficiently solve the straggler problem in Sync FL, yielding faster convergence. In summary, under highly skewed data distribution, i.e., $\theta \to 0$, TT-Fed can improve the converged test accuracy by up to 12.5\% and 5\% compared to FedAsync and FedAT.

\subsection{Communication Overhead Comparisons}

Fig. \ref{Fig_sec5_Communication_Round_Comparisons} plots the communication overhead of the four FL algorithms under parameter settings $\theta \to 0,~\eta=0$, where the number of communication rounds required to reach the given target accuracy is shown. The CPU frequency of each user is randomly drawn from the range of $1 \sim 5 ~ \rm{GHz}$, and the global aggregation round duration of TT-Fed is $\Delta T = 0.6T$. For fair comparison, each tier in FedAT contains the same users as that in TT-Fed. From Fig. \ref{Fig_sec5_Communication_Round_Comparisons}, we see that FedAsync requires larger number of communication rounds to reach a given target accuracy compared to its Sync counterpart (FedAvg). This is because in FedAsync the server has to frequently transmit the global model to whichever users ready for the next local updating round, whereas, in FedAvg users are synchronized at each global aggregation round, and model broadcasting can boost communication efficiency. Moreover, FedAsync suffers from severe computation bias which increases the number of communication rounds required to go from a lower target accuracy (i.e., 70\%) to a higher one (i.e., 80\%). Fig. \ref{Fig_sec5_Communication_Round_Comparisons} also shows that with the same tier partitioning, TT-Fed achieves improved communication efficiency compared to the event-triggered FedAT algorithm. This is because users in different tiers are aligned in TT-Fed to receive the global model via broadcasting, while in FedAT users in different tiers obtain the global model asynchronously.

\subsection{Impact of Training Round Partitioning}
\begin{figure}[t!]
\begin{minipage}[t]{0.45\linewidth}
    \includegraphics[scale=0.55]{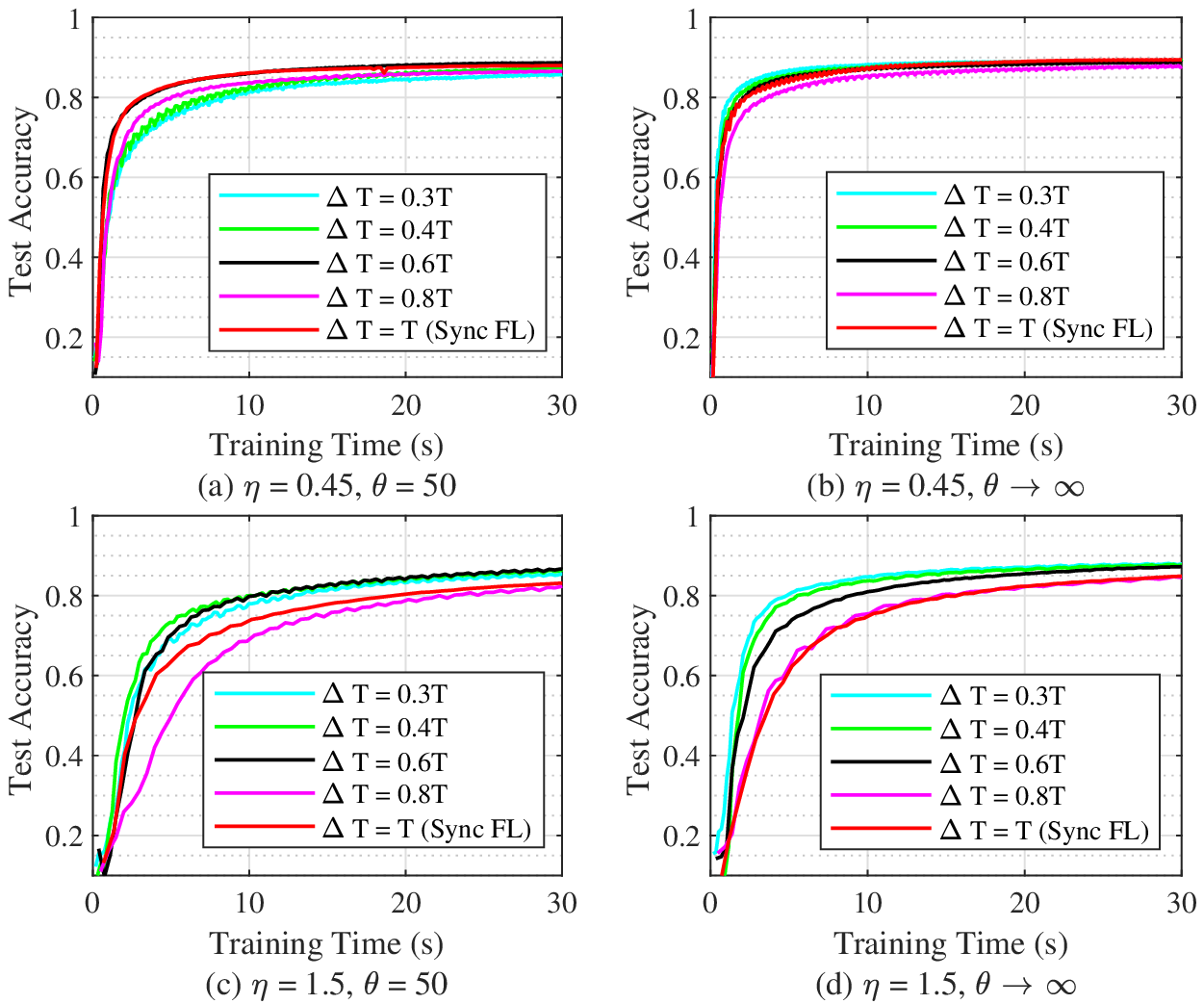}
    \caption{Test accuracy of TT-Fed under different values of $\Delta T$.}
    \label{Fig_sec5_Round_Length_comparisons}
\end{minipage}%
    \hfill%
\begin{minipage}[t]{0.45\linewidth}
    \includegraphics[scale=0.55]{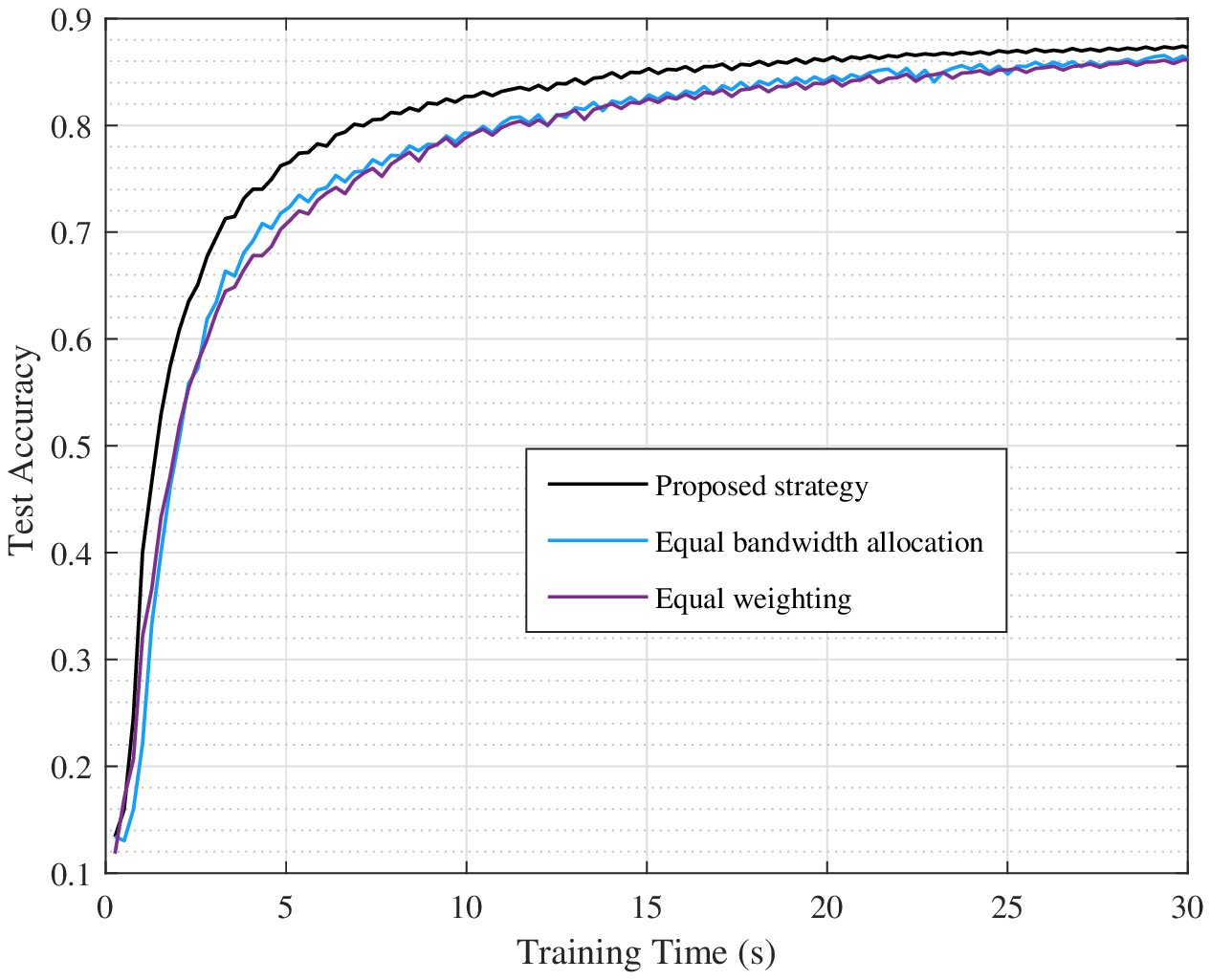}
    \caption{Performances of TT-Fed under different bandwidth allocation and user selection strategies.}
    \label{Fig_sec5_Schedluing_policy_comparisons}
\end{minipage} 
\end{figure}

In Fig. \ref{Fig_sec5_Round_Length_comparisons}, we examine how global aggregation round duration of TT-Fed affect its test accuracy under various data distributions. The simulations plot the test accuracies with 4 tiers ($\Delta T = 0.3T$), 3 tiers ($\Delta T=0.4T$), 2 tiers ($\Delta T =0.6T $, $0.8T$) and single tier ($\Delta T =T$ equals to Sync FL.). In Fig. \ref{Fig_sec5_Round_Length_comparisons} (a), the global aggregation round duration settings $\Delta T =0.6T$ and $\Delta T =T$ both yields the highest test accuracy among all considered settings, while $\Delta T =0.3T$ yields the worst performance. In this case, the computation delay is similar for each user but data class distribution is highly skewed. Therefore, to ensure  a good estimation of the global gradient, we need more users in a tier to guarantee that the data diversity is similar enough to the global data distribution. This explains why partitioning users into more tiers will suffer more from the skewed data class distribution. Second, with the same number of user tiers, the performance of $\Delta T = 0.8T$ is degraded compared to $\Delta T = 0.6T$ in Fig. \ref{Fig_sec5_Round_Length_comparisons} (a). This is because the global aggregation round length of $\Delta T = 0.6T$ is smaller than that of $\Delta T = 0.8T$ and the user are more evenly partitioned. In Fig. \ref{Fig_sec5_Round_Length_comparisons} (d), when $\eta=1.5$ and $ \theta \to \infty$, partitioning users into more tiers is more beneficial (i.e., $\Delta T = 0.3T$). This is because a tier with fewer users contains enough information to generate a good estimation of the global gradient.

For $\eta=0.45$, $\theta =50$ and $\eta=1.5$, $\theta=50$ in Fig. \ref{Fig_sec5_Round_Length_comparisons} (a) (c), we see that partitioning users into more tiers is not always beneficial. This is because the data class distribution is highly skewed in these two scenarios, dividing users into many small tiers may yield biased gradient estimation for global loss function. However, when the differences in computation delay is large and data class distribution is similar, dividing users into smaller tiers will be beneficial ($\eta=1.5$, $\theta \to \infty$ in Fig. \ref{Fig_sec5_Round_Length_comparisons} (d)). Interestingly, when the computation delay and data class distribution of users are similar ($\eta=0.45$, $\theta \to \infty$ in Fig. \ref{Fig_sec5_Round_Length_comparisons} (b)), little gain can be obtained with more user tiers. In summary, the optimal global aggregation round duration setting in TT-Fed is the smallest one that generates unbiased estimate to the global information.

\subsection{Scheduling Policy Comparisons}
Fig. \ref{Fig_sec5_Schedluing_policy_comparisons} plots the test accuracies of TT-Fed under different resource allocation and user selection policies, with $\Delta T = 0.6T$, $\eta = 1$ and $\theta = 100$. For comparison purpose, the two used baselines are: a) equal bandwidth allocation with optimal user selection, and b) optimal bandwidth allocation and user selection under equal weighting (i.e., $\alpha_m^k = \frac{1}{M}$). From Fig. \ref{Fig_sec5_Schedluing_policy_comparisons}, we observe that our proposed resource allocation and user selection strategy outperforms both baselines a) and b) in convergence speed and accuracy. Compared to baseline a), the proposed bandwidth allocation algorithm can adapt to channel conditions of different users, so that the success probability of uplink model transmission is enhanced. Compared to baseline b), our proposed scheme can alleviate the computation bias imposed by frequent updates from fast tiers.

\section{Conclusion}
In this paper, we proposed a multi-tier time-triggered FL algorithm (TT-Fed) for resource-limited and unreliable wireless networks. We formulated a joint user selection and bandwidth allocation problem with the aim of training loss minimization. To decompose this problem into tractable sub-problems, we provided a detailed convergence analysis and derived an analytical convergence upper bound for our proposed TT-Fed algorithm. Moreover, a sufficient condition was given to guarantee the convergence of TT-Fed. Finally, we proposed an online search algorithm to solve the formulated problem. Our simulation results showed that the proposed TT-Fed algorithm can achieve fast convergence with robust performance under highly non-IID data, while substantially reducing the communication overhead.
 
\appendices

\section{Proof of Theorem 1}\label{Convergence_Analysis}
First, we introduce a lemma which will be useful in the following derivation process.
\begin{lemma}
For a $\mu$-strong convex function $F$ with optimum solution $x^*$, the following inequality holds for $\forall x \in \mathrm{dom}{(F)}$
\begin{align}
2 \mu [F(x) - F(x^*)] \le ||\nabla F(x)||^2 .  \label{lemma1_eq}
\end{align}
\label{cvx_ineq}
\end{lemma}
\vspace{-1cm}
\begin{proof}
Given $x$, we can construct the following function
\begin{align}
g(y) = F(x) + (y-x)^\intercal \nabla F(x)  + \frac{\mu}{2}||y - x||^2.
\end{align}

Following the definition of $\mu$-strong convexity in \eqref{assum:cvx}, we have $F(x^*) \ge g(x^*)$. As $\nicefrac{\mu}{2} \ge 0$, $g(y)$ is a convex quadratic function of $y$. Therefore, the minimum of $g(y)$ is obtained at point  $\nabla g(y) = 0$ with value $F(x) - \nicefrac{||\nabla F(x)||^2}{2\mu}$. Thus
\begin{align}
F(x^*) \ge F(x) - \frac{||\nabla F(x)||^2}{2\mu}.
\label{lemma1_der1}
\end{align}

Rearranging terms in \eqref{lemma1_der1} yields the result.
\end{proof}

We define $\nabla f( w^{k-1}_{\text{G}};u)  \buildrel \Delta \over =  \sum\limits_{i=1}^{D_u} \nabla f( w^{k-1}_{\text{G}};x_{u,i},y_{u,i})$, to serve the conciseness of the following derivation. Following the global aggregation in \eqref{TT_global}, we have 
\begin{align}
& F(w^k_{\text{G}}) - F(w^{k-1}_{\text{G}})  \nonumber \\ 
&  =  F \Big[ \sum _{m=1}^M \big( 1-  \mathds{1} \{ k \bmod m =0 \} \big)  \alpha ^k_m w^{k-1}_{\text{G}} + \sum _{m=1}^M \mathds{1} \{ k \bmod m =0 \} \alpha^k_m  w ^k_{\text{I},m} \Big] - F(w^{k-1}_{\text{G}}).
\end{align}
Based on the convexity of $F$ and the Jensen's inequality, we have
\begin{align}
& F(w^k_{\text{G}}) - F(w^{k-1}_{\text{G}})  \nonumber \\ 
&  \le  \sum _{m=1}^M \Big( 1- \mathds{1} \{ k \bmod m =0 \} \Big) \alpha ^k_m F(w^{k-1}_{\text{G}}) +  \sum _{m=1}^M \mathds{1} \{ k \bmod m =0 \} \alpha^k_m F(w ^k_{\text{I},m}) - F(w^{k-1}_{\text{G}}) \nonumber \\ 
& = \sum _{m=1}^M \mathds{1} \{ k \bmod m =0 \} \alpha^k_m \big[ F(w ^k_{\text{I},m}) - F(w^{k-1}_{\text{G}}) \big].    \label{der1}
\end{align}

Now, we derive the term $F(w^k_{\text{I},m}) - F(w^{k-1}_{\text{G}})$ in \eqref{der1}. Let us denote $\kappa = \nicefrac{\sum\limits_{u \in {\cal{S}}_{m,\text{S}}} \nabla f(w^{k-m}_{\text{G}};u) }{D_{m,\text{S}}}$.
Based on $L$-smoothness property of $F$ in \eqref{assum:smooth}, and updating rules \eqref{TT_intra}, \eqref{TT_local}, we have
\begin{align}
F(w^k_{\text{I},m}) - F(w^{k-1}_{\text{G}}) &  \le  \big( w^k_{\text{I},m} - w^{k-1}_{\text{G}} \big)^\intercal \nabla F(w^{k-1}_{\text{G}})  + \frac{L}{2}||w^k_{\text{I},m} - w^{k-1}_{\text{G}}||^2  \nonumber \\
&  =  \big( w^{k-m}_{\text{G}} - w^{k-1}_{\text{G}} - \lambda \kappa \big) ^\intercal \nabla F(w^{k-1}_{\text{G}}) + \frac{L}{2} \left\|  {w^{k-m}_{\text{G}} - w^{k-1}_{\text{G}} }- \lambda \kappa \right\|^2.\label{der2-1}
\end{align}
Noticing $\|{w^{k-m}_{\text{G}} - w^{k-1}_{\text{G}} } - \lambda \kappa \|^2 \le 2 (\| {w^{k-m}_{\text{G}} - w^{k-1}_{\text{G}} } \|^2 +\lambda ^2 \| \kappa \|^2)$ in the last term of \eqref{der2-1}, and based on assumptions \eqref{assum:global_1}, \eqref{assum:global_2}, we have
\begin{align}
 F(w^k_{\text{I},m}) - F(w^{k-1}_{\text{G}}) & \le  \delta \left\| \nabla F(w^{k-1}_{\text{G}}) \right\|^2 - \lambda \kappa ^ \intercal \nabla F(w^{k-1}_{\text{G}})  +  L\varepsilon^2  + L \lambda ^2 \left\| \kappa \right\|^2 \nonumber \\
& = \delta \left\| \nabla F(w^{k-1}_{\text{G}}) \right\|^2 - \lambda \left[\nabla F(w^{k-1}_{\text{G}}) - (\nabla F(w^{k-1}_{\text{G}}) - \kappa)  \right] ^ \intercal \nabla F(w^{k-1}_{\text{G}})  \nonumber \\ 
& \hspace{0.6cm} +  L\varepsilon^2  + L \lambda ^2 \left\|  \left[\nabla F(w^{k-1}_{\text{G}}) - (\nabla F(w^{k-1}_{\text{G}}) - \kappa)  \right] \right\|^2.
\label{der2}
\end{align}
Setting $\lambda = \frac{1}{2L}$ in \eqref{der2} and rearranging terms, we have
\begin{align}
F(w^k_{\text{I},m}) - F(w^{k-1}_{\text{G}}) & =  L\varepsilon^2 + (\delta -\frac{1}{4L}) \| \nabla F(w^{k-1}_{\text{G}})  \|^2 + \frac{1}{4L} \|\nabla F(w^{k-1}_{\text{G}}) - \kappa  \|^2. \label{der3}
\end{align} 

Now, we derive the last term in \eqref{der3}. Following the definition of $F$, we have 
\begin{align}
  \nabla F(w^{k-1}_{\text{G}})   
=  \nicefrac{ \sum\limits_{j=1}^{M} \sum\limits_{u \in {\cal{S}}_j} \nabla f( w^{k-1}_{\text{G}};u) }{D} .  \label{der3-1}
\end{align}
Let us denote $\Omega = \nicefrac{\sum_{j=1,j \ne m }^{M} \sum\limits_{u \in {\cal{S}}_j} \nabla f( w^{k-1}_{\text{G}};u)}{D}$.
Then, the following inequality holds
\begin{align}
\nabla F(w^{k-1}_{\text{G}})  \le \Omega  + \nicefrac{\sum\limits_{u \in {\cal{S}}_m} \nabla f( w^{k-1}_{\text{G}};u)}{D_m}. \label{scale1}
\end{align}
By plugging into \eqref{scale1} and the expression of $\kappa$, the last term in \eqref{der3} can be scaled to
\begin{align}
\| \nabla F(w^{k-1}_{\text{G}}) -  \kappa  \|^2  & \le \Big\| \Omega + \nicefrac{\sum\limits_{u \in {\cal{S}}_m} \nabla f( w^{k-1}_{\text{G}};u)}{D_m}  -  \nicefrac{\sum\limits_{u \in {\cal{S}}_{m,\text{S}}} \nabla f( w^{k-m}_{\text{G}};u)}{D_{m,\text{S}}} \Big\|^2 \nonumber \\
&  \mathop = ^{(a)}  \Big\| \Omega + \nicefrac{\sum\limits_{u \in {\cal{S}}_{m,\text{S}}}  [ \nabla f( w^{k-1}_{\text{G}};u) - \nabla f( w^{k-m}_{\text{G}};u)  ]}{D_m }  + \nicefrac{  \sum\limits_{u \in {\cal{S}}_{m,\text{F}}} \nabla f( w^{k-1}_{\text{G}};u)}{D_m} \nonumber \\ 
& \hspace{0.4cm} -  \nicefrac{ \big( D_{m,\text{F}} \sum\limits_{u \in {\cal{S}}_{m,\text{S}}} \nabla f( w^{k-m}_{\text{G}};u) \big)}{\big( D_m D_{m,\text{S}} \big)}  \Big\|^2, \label{der3-2}
\end{align}
where $(a)$ comes from plugging into the expressions of $D_{m,\text{S}}$ and $D_{m,\text{F}}$, and noticing the fact  ${\cal{S}}_{m,\text{F}}  = {\cal{S}}_m \backslash {\cal{S}}_{m,\text{S}}$. Substituting the expression of $\Omega$ and applying triangle inequality to \eqref{der3-2}, we yield
\begin{align} 
\| \nabla F(w^{k-1}_{\text{G}}) -  \kappa \|^2  & \le  \bigg[ \nicefrac{\sum\limits_{j=1,j \ne m }^{M} \sum\limits_{u \in {\cal{S}}_k}   \| \nabla f( w^{k-1}_{\text{G}};u)  \|}{D}  + \nicefrac{ \sum\limits_{u \in {\cal{S}}_{m,\text{S}}} \big(   \| \nabla f( w^{k-1}_{\text{G}};u) - \nabla f( w^{k-m}_{\text{G}};u)  \| \big)  }{D_m} \nonumber \\
& \hspace{0.4cm} + \nicefrac{ \sum\limits_{u \in {\cal{S}}_{m,\text{F}}}  \| \nabla f( w^{k-1}_{\text{G}};u) \|}{D_m }  +   \nicefrac{\big( D_{m,\text{F}} \sum\limits_{u \in {\cal{S}}_{m,\text{S}}}  \| \nabla f( w^{k-m}_{\text{G}};u)  \| \big)}{\big(D_m D_{m,\text{S}} \big)} \bigg]^2. \label{der3-3} 
\end{align}

Applying \eqref{assum:local_2} to the second term of \eqref{der3-3}, and \eqref{assum:local_1} to the last term of \eqref{der3-3}, then according to \eqref{assum:local_global}, we have
\begin{align}
& \| \nabla F(w^{k-1}_{\text{G}}) -  \kappa  \|^2  \nonumber \\
& \le   \Big[\big( \frac{D-D_m}{D} \big) \sqrt { \chi +\nu  \| \nabla F( w^{k-1}_{\text{G}}) \|^2 }  + \frac{\phi D_{m,\text{S}} }{D_m}  +  \big( \frac{ (1 + \beta) D_{m,\text{F}} }{D_m } \big) \sqrt { \chi +\nu \| \nabla F( w^{k-1}_{\text{G}}) \|^2 }  \Big]^2.  \label{der3-4}
\end{align}
Applying the Cauchy-Schwartz inequality $(x+y+z)^2 \le 3(x^2+y^2+z^2)$ to \eqref{der3-4}, and noticing $D_{m,\text{S}},\, D_{m,\text{F}} \le D_m$, we have
\begin{align}
&  \| \nabla F(w^{k-1}_{\text{G}}) -  \kappa   \|^2  \le  3 \Big\{ \phi^2 +\big( 1+ \frac{ (1 + \beta)^2 D_{m,\text{F}} }{D_m } \big) \big( \chi +\nu \| \nabla F( w^{k-1}_{\text{G}})  \|^2 \big)  \Big\}.  \label{der4}
\end{align}

Substituting \eqref{der4} and \eqref{der3} back to \eqref{der1} and rearranging terms, we have
\begin{align}
F(w^k_{\text{G}}) - F(w^{k-1}_{\text{G}}) & \le \sum _{m=1}^M \mathds{1} \{ k \bmod m =0 \} \alpha^k_m \big[ \Delta_1(m)  -  \frac{1}{4L} \Delta_2(m)||\nabla F(w^{k-1}_{\text{G}})||^2 \big],  \label{der5-1}
\end{align}
where 
\begin{align}
\Delta_1(m) & = L\varepsilon^2 + \frac{3}{4L}\big[ \phi^2 + \chi \big( 1+ \frac{(1+\beta)^2 D_{m,\text{F}}}{D_m} \big) \big] ,\\
\Delta_2(m) & = 1-4\delta L -3\nu\big[ 1+ \frac{(1+\beta)^2 D_{m,\text{F}}}{D_m} \big]. 
\end{align}

According to median value theorem, there exists a positive number $\xi \in (0,M)$  that makes the following equality hold
\begin{align}
&  \sum _{m=1}^M \mathds{1} \{ k \bmod m =0 \} \alpha^k_m \big[ \Delta_1(m) -  \frac{ \Delta_2(m)}{4L}||\nabla F(w^{k-1}_{\text{G}})||^2 \big] = \xi \big[ \Delta_1-  \frac{ \Delta_2(m)}{4L} ||\nabla F(w^{k-1}_{\text{G}})||^2 \big], \label{der5-2}
\end{align}
where $\Delta_i = \frac{1}{M}\sum_{m=1}^M \Delta_i (m), ~i=\{1,2\}$. 

Substituting \eqref{der5-2}, \eqref{lemma1_eq} into \eqref{der5-1}, and subtracting $F(w^*_{\text{G}})-F(w^{k-1}_{\text{G}})$ from both sides, we have
\begin{align}
F(w^k_{\text{G}})-F(w^*_{\text{G}})  \le  \big( 1-\frac{\mu\xi}{2L}\Delta_2 \big)\big[F(w^{k-1}_{\text{G}}) - F(w^*_{\text{G}})\big] + \xi\Delta_1. \label{der6}
\end{align}
Recursively using \eqref{der6} and taking the expectation with respect to successful transmission probability on both sides, we have
\begin{align}
\mathbb{E} \{ F(w^k_{\text{G}})-F(w^*_{\text{G}}) \}  \le \mathbb{E} \Big\{ \big( 1-\frac{\mu\xi}{2L}\Delta_2 \big)^k [F(w^{0}_{\text{G}}) - F(w^*_{\text{G}})] +  \frac{2\Delta_1 L}{\mu \Delta_2} \big[1-\big( 1- \frac{\mu\xi}{2L}\Delta_2 \big)^k \big] \Big\} , \label{der7}
\end{align} 
where $w^0_{\text{G}}$ is the initial global model. Changing $k$ to $K$ directly yields the result.

\section{Proof of Theorem 2}\label{bandwidth_proof}
Given the user selection strategy, the second order derivative of the optimization goal in \eqref{Pg_goal} with respect to bandwidth allocation variable $b_u^k$ is
\begin{align}
\sum _{m=1}^M \sum _{u \in {\cal{S}}_m} \mathds{1}\{k\bmod m =0\} \alpha_m^k a^k_u  \times D_u \big( \frac{\gamma_{\rm{th}} N_0}{P l(d_u^k)}\big)^2 e^{-\frac{\gamma_{\rm{th}} N_0 b_u^k }{P l(d_u^k)}}  \ge 0, \label{pf2_der1}
\end{align}
which indicates that the optimization goal is a convex function. Since the maximum of a convex function is obtained at the constraint boundary \cite{boyd2004convex}, under unlimited bandwidth budget, the optimum bandwidth allocation for a selected user is obtained at the boundary, which is
\begin{align}
\tau ^k_{u,\text{cm}} + \tau ^k_{u,\text{cp}} = m \Delta T,  \; \forall u \in {\cal{S}}_m. \label{pf2_der2}
\end{align} 

Substituting \eqref{delay_cm} and \eqref{rate} into \eqref{pf2_der2} and rearranging terms, we have
\begin{align}
b_u^k \log _2 \big( 1+ \frac{P \| g_u^k \|^2}{N_0 b_u^k} \big)  = \frac{Z}{m \Delta T - \tau ^k_{u,\text{cp}}}. \label{pf2_der3}
\end{align}

To serve the conciseness of the following derivation, we denote $A = \frac{Z}{m \Delta T - \tau ^k_{u,\text{cp}}}$, $C = \frac{P \| g_u^k \|^2}{N_0}$, $\nicefrac{1}{b_u^k} = x$. Rearranging terms in \eqref{pf2_der3}, we yield
\begin{align}
2^{Ax} = 1+Cx. \label{pf2_der4}
\end{align}
Multiplying $\frac{A}{C} 2^{-\left( Ax + \frac{A}{C} \right)}$ to both sides of \eqref{pf2_der4}, we have
\begin{align}
\frac{A}{C} {2^{-\frac{A}{C}}} = \big(\frac{A}{C} + Ax \big) 2^{-( \frac{A}{C} + Ax )}. \label{pf2_der5}
\end{align}
Let us denote $t = \nicefrac{A}{C} + Ax$. Multiplying $-\ln 2$ to both sides of \eqref{pf2_der5}, we yield
\begin{align}
- \frac{A}{C} {2^{-\frac{A}{C}}} \ln 2 = (-t\ln 2) e^{(-t\ln 2)}. \label{pf2_der6}
\end{align}
Following the definition of Lambert-$W$ function \cite{Corless96onthe}, we have
\begin{align}
-t\ln 2 = W \big( - \frac{A}{C} {2^{-\frac{A}{C}}} \ln 2 \big). \label{pf2_der7}
\end{align}
Substituting $t = \nicefrac{A}{C} + \nicefrac{A}{b_u^k}$ into \eqref{pf2_der7} and rearranging terms, we have
\begin{align}
b_u^k = \frac{-A \ln2}{W \big( - \ln 2 \frac{A}{C}  {e^{-\frac{A}{C} \ln2}}  \big) + \frac{A}{C} \ln2}, \label{pf2_der8}
\end{align}
where the Lambert-$W$ function will have two solutions on the main branch and on the $W_{-1}$ branch, respectively, since $- \ln 2 \frac{A}{C}  {e^{-\frac{A}{C} \ln2}}$ is bounded by $(0,-\nicefrac{1}{e}]$. We exclude the one on the main branch $W_0$ which makes the denominator in \eqref{pf2_der8} to be $0$. So, the optimum result is obtained on the $W_{-1}$ branch. Finally, substituting $A,~C,~x,~t$ into \eqref{pf2_der8} directly yields the result.

\bibliographystyle{IEEEtran}
\bibliography{IEEEabrv,TT_bib}

\balance
\end{document}